%% file: main.tex
\documentclass[10pt,twocolumn,letterpaper]{article}

\input{packages.tex}
\begin{document} 

\input{sections/title.tex}
\input{sections/abstract.tex}

\section{Introduction} \label{sec:intro}
\input{sections/introduction.tex}

\section{Related Work}
\label{sec:related}
\input{sections/related_work.tex}

\section{Background: Single-Photon Imaging Model}
\label{sec:background}
\input{sections/background.tex}

\section{Projections of the Photon-Cube}
\label{sec:projections}
\input{sections/projections.tex}

\section{Emulating Cameras from Photon-Cubes}
\label{sec:emulating}
\input{sections/emulating.tex}

\section{Hardware and Experimental Results}
\label{sec:experiments}
\input{sections/experiments.tex}

\section{Discussion and Future Outlook}
\label{sec:discussion}
\input{sections/discussion.tex}

{\small

\input{main.bbl}
}

\onecolumn 
\setcounter{section}{0}
\setcounter{equation}{0}
\setcounter{figure}{0}
\setcounter{table}{0}
\setcounter{page}{1}
\makeatletter

\renewcommand{\figurename}{Supplementary Figure}
\captionsetup[figure]{name={Supplementary Figure}}
\renewcommand{\theequation}{S\arabic{equation}}
\crefname{supplementaryfigure}{Suppl. Fig.}{Suppl. Figs.}
\Crefname{supplementaryfigure}{Supplementary Figure}{Supplementary Figures}
\crefalias{figure}{supplementaryfigure}%
\renewcommand*{\thefootnote}{$\dagger$}

\clearpage
\begin{center}
\textbf{\large Supplementary Material for ``SoDaCam: Software-defined Cameras via Single-Photon Imaging''}
\end{center}
\input{supplementary.tex}

\end{document}

%% file: packages.tex
\usepackage{iccv}
\usepackage{times}
\usepackage{epsfig}
\usepackage{graphicx}
\usepackage{amsmath}
\usepackage{amssymb}
\usepackage[accsupp]{axessibility}  

\usepackage{booktabs}
\usepackage{comment}
\usepackage{tabularx}
\newcolumntype{L}{>{\raggedright\arraybackslash}X}

\usepackage{caption}
\usepackage{color}
\usepackage{mathtools}
\usepackage{multirow}
\usepackage{multicol}
\usepackage{enumitem}
\usepackage{siunitx}
\usepackage{makecell}

\usepackage[numbers,sort,compress]{natbib}

\usepackage{hyperref}
\hypersetup{pagebackref=true,breaklinks=true,letterpaper=true,colorlinks,bookmarks=false}
\iccvfinalcopy 

\def\iccvPaperID{7164} 

\makeatletter

\usepackage[capitalize]{cleveref}
\crefname{section}{Sec.}{Secs.}
\Crefname{section}{Section}{Sections}
\Crefname{table}{Table}{Tables}
\crefname{table}{Tab.}{Tabs.}

\makeatother

\input{math_macros.tex}

\DeclareCaptionLabelFormat{andfigure}{#1~#2  \&  \figurename~\thefigure}
\DeclareCaptionLabelFormat{andtable}{#1~#2  \&  \tablename~\thetable}

\newcommand{\tightcaption}[1]{\vspace{-0.05in}\caption{#1}\vspace{-0.0in}}

\usepackage{algorithm}
\usepackage{algpseudocode}
\usepackage[frozencache,cachedir=.,newfloat]{minted}

\newcommand*{\crefex}[1]{%
Suppl. \cref{#1}}

\newenvironment{code}{\captionsetup{type=listing}}{}

\DeclareCaptionLabelFormat{andfigure}{#1~#2  \&  \figurename~\thefigure}

\DeclareCaptionLabelFormat{andtable}{#1~#2  \&  \tablename~\thetable}

%% file: math_macros.tex
\newcommand{\vx}{\mathbf{x}}

\newcommand{\cx}{\mathcal{X}}

\newcommand{\cI}{\mathcal{I}}

\newcommand{\vr}{\mathbf{r}}

\newcommand{\rr}{\mathbb{R}}

\newcommand{\uvec}[1]{\mathbf{\hat{#1}}}

\newcommand{\ceil}[1]{\left\lceil#1\right\rceil} 
\newcommand{\abs}[1]{\left\lvert#1\right\rvert}

\newcommand{\Prob}[1]{\text{Pr}\left\{#1 \right\}}

\newcommand{\tildeNice}{{\raise.17ex\hbox{$\scriptstyle\sim$}}}
\newcommand{\approxNice}{{\raise.15ex\hbox{$\scriptstyle\approx$}}}
\newcommand{\ggNice}{{\raise.15ex\hbox{$\scriptstyle >> $}}}

%% file: sections/title.tex
\title{SoDaCam: Software-defined Cameras via Single-Photon Imaging}

\author{
  Varun Sundar$^\dagger$\\
  {\tt\small vsundar4@cs.wisc.edu}\and
  Andrei Ardelean$^\ddagger$\\
  {\tt\small a.ardelean@epfl.ch}\and
  Tristan Swedish$^\S$\\
  {\tt\small tristan@ubicept.com}\and
  Claudio Bruschini$^\ddagger$ \quad Edoardo Charbon$^\ddagger$\\
  {\tt\small \{claudio.bruschini,}
  {\tt\small edoardo.charbon\}@epfl.ch}\and
  Mohit Gupta$^{\dagger,\S}$\\
  {\tt\small mohitg@cs.wisc.edu}\and
  {$^\dagger$University of Wisconsin-Madison}\quad
  {$^\ddagger$École polytechnique fédérale de Lausanne}\quad
  {$^\S$Ubicept}}

\maketitle
\renewcommand*{\thefootnote}{$\ast$}
\setcounter{footnote}{1}
\footnotetext{Corresponding author: Varun Sundar. This research was supported in parts by NSF CAREER award 1943149, NSF award CNS-2107060, and the Swiss National Science Foundation grant 200021\_166289. We also thank Paul Mos for providing access to SwissSPAD2 acquisition software.}

\renewcommand*{\thefootnote}{\arabic{footnote}}
\setcounter{footnote}{0}

%% file: sections/abstract.tex
\begin{abstract}
Reinterpretable cameras are defined by their post-processing capabilities that exceed traditional imaging. We present ``SoDaCam'' that provides reinterpretable cameras at the granularity of photons, from photon-cubes acquired by single-photon devices. Photon-cubes represent the spatio-temporal detections of photons as a sequence of binary frames, at frame-rates as high as 100 kHz. We show that simple transformations of the photon-cube, or photon-cube projections, provide the functionality of numerous imaging systems including: exposure bracketing, flutter shutter cameras, video compressive systems, event cameras, and even cameras that move during exposure. Our photon-cube projections offer the flexibility of being software-defined constructs that are only limited by what is computable, and shot-noise. We exploit this flexibility to provide new capabilities for the emulated cameras. As an added benefit, our projections provide camera-dependent compression of photon-cubes, which we demonstrate using an implementation of our projections on a novel compute architecture that is designed for single-photon imaging.
\end{abstract}

%% file: sections/introduction.tex
Throughout the history of imaging, sensing technologies and the corresponding processing have developed hand-in-hand. In fact, sensing technologies have, to some extent, defined the scope of processing captured data. In the film era, instances of such processing included dodging and burning. The advent of digital cameras provided processing at the granularity of pixels and paved the way for modern computer vision.  Light field cameras~\cite{levoy1996light,zhang2017light}, by sampling the plenoptic function \cite{adelson1991plenoptic}, allowed post-capture processing at the granularity of light rays, enabling novel functionalities such as refocusing photos after-capture. The logical limit of post-capture processing, given the fundamental quantization of light, would be at the level of individual photons. What would imaging look like if we could perform computational processing on individual photons? 

In this work, we show that photon data captured by a new class of single-photon detectors, called single-photon avalanche diodes (SPADs), makes it possible to emulate a wide range of imaging modalities such as exposure bracketing~\cite{debevec2008recovering}, video compressive systems~\cite{CACTI_Llull:13,P2C2} and event cameras~\cite{ATIS_2011,serrano2013128}. A user then has the flexibility to choose one (or even multiple) of these functionalities \textit{post-capture} (\cref{fig:first-fig} \textit{(top)}).
SPAD arrays can operate as extremely high frame-rate photon detectors (\tildeNice $100$~kHz), producing a temporal sequence of binary frames called a photon-cube \cite{Fossum:11}. 
We show that computing \textit{photon-cube projections}, which are simple linear and shift operations, can reinterpret the photon-cube to achieve novel post-capture imaging functionalities in a software-defined manner (\cref{fig:first-fig} \textit{(middle)}). 

As case studies, we emulate three distinct imagers: high-speed video compressive imaging; event cameras which respond to dynamic scene content; and motion projections which emulate sensor motion, without any real camera movement. \cref{fig:first-fig} \textit{(bottom)} shows the outputs of these cameras that are derived from the same photon-cube.

\input{figures/first_fig.tex}

\paragraph{Computing photon-cube projections.} One way to obtain photon-cube projections is to read the entire photon-cube off the SPAD array and then perform relevant computations off-chip; we adopt this strategy for our experiments in \cref{sec:swiss-spad2,sec:comp-real-hardware}. 
While reasonable for certain applications, reading out photon-cubes requires an exorbitant data bandwidth, which can be up to $100$ Gbps for a $1$ MPixel array---well beyond the capacity of existing data peripherals. Such readout considerations will become center stage as large-format SPAD arrays are fabricated~\cite{morimoto_megapixel_2020,Morimoto_cannon}.

An alternative is to avoid transferring the entire photon-cube by computing projections near sensor.  
As a proof-of-concept, we implement photon-cube projections on UltraPhase \cite{ardelean2023computational}, a recently-developed programmable SPAD imager with independent processing cores that have dedicated RAM and instruction memory. We show, in \cref{sec:ultraphase}, that computing projections on-chip greatly reduces sensor readout and, as a consequence, power consumption.

\paragraph{Implications: Toward a photon-level software-defined camera.}The photon-cube projections introduced in this paper are computational constructs that provide a realization of \textit{software-defined cameras} or \textit{SoDaCam}. Being software-defined, SoDaCam can emulate multiple cameras simultaneously without additional hardware complexity.
SoDaCam, by going beyond baked-in hardware choices, unlocks hitherto unseen capabilities---such as $2000$ FPS video from $25$ Hz readout (\cref{fig:comp-video}); event imaging in very low-light conditions (\cref{fig:prophesse-comp});  and motion stacks, which are a stack of images wherein each image, objects only in certain velocity ranges appear sharp (\cref{fig:motion-stack}).

\paragraph{Limitations.}The SPAD array \cite{ulku512512SPAD2019} used in this work has a relatively low spatial resolution ($512 \times 256$), and a low fill-factor (\tildeNice$10$\%) owing to the lack of microlenses in the prototype used. Similarly, the near-sensor processor that we use has limited capabilities compared to off-chip processors. However, with rapid progress in the development of single-photon cameras \cite{morimoto_megapixel_2020,Morimoto_cannon} and increasing interest in near-sensor processors, we anticipate that many of these shortcomings will be addressed in the upcoming years.

%% file: figures/first_fig.tex
\begin{figure*}[t]
    \centering
    \includegraphics[width=\textwidth]{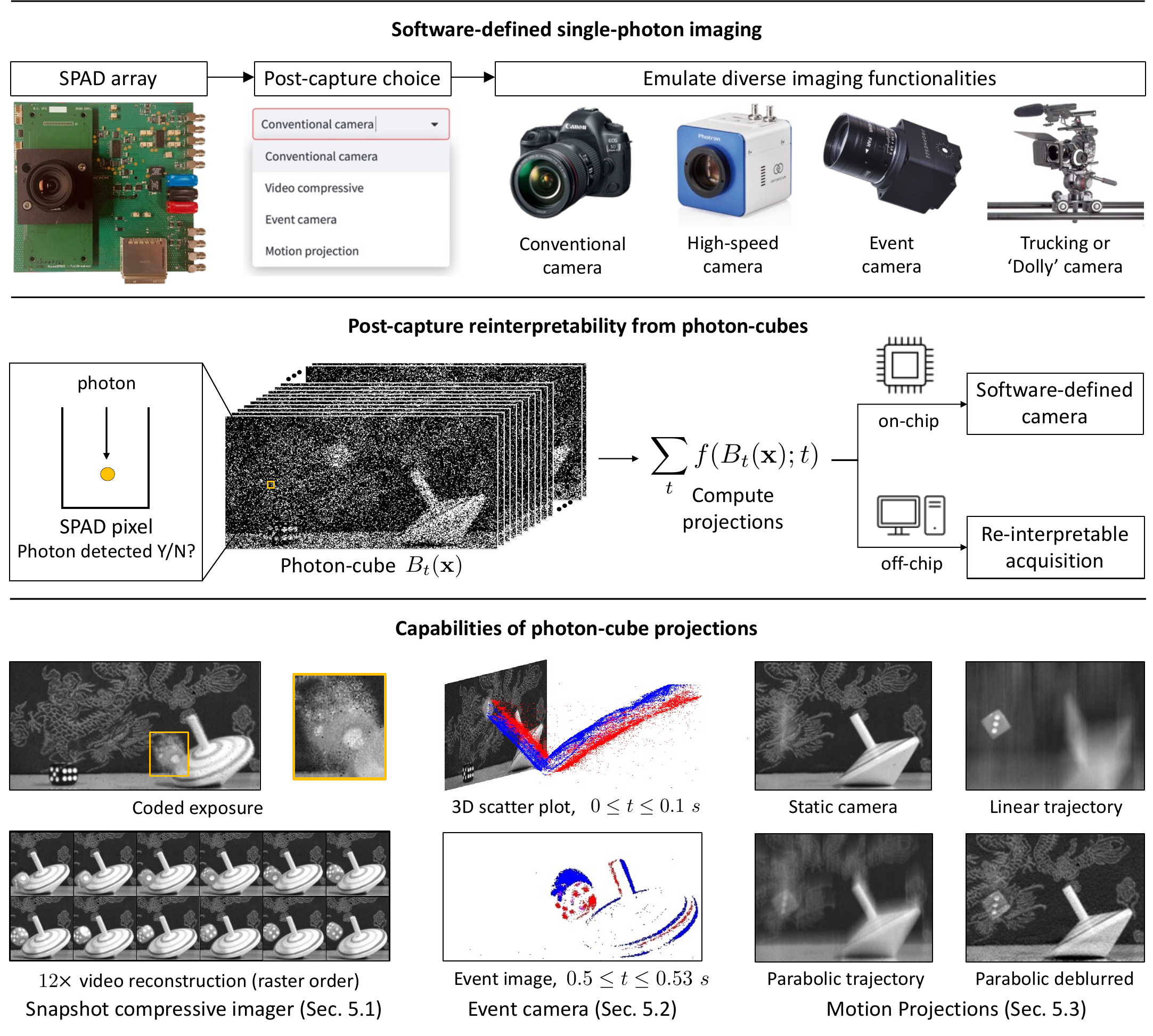}  
    \vspace{-0.2in}
    \tightcaption{\textit{(top)} \textbf{SoDaCam} can emulate a variety of cameras from the photon-cubes acquired by single-photon devices. \textit{(middle)} Photon-cubes represent the spatio-temporal detection of photons as a sequence of binary frames. 
    Projections of the photon-cube, when computed either on or off-chip, result in reinterpretable and software-defined cameras.
    We demonstrate the versatility of photon-cube projections on a \textbf{real dynamic scene}: a die falls on a table, bounces, spins in the air, and later ricochets off a nearby toy top. \textit{(bottom)} The cameras emulated by our photon-cube projections can produce a $12\times$ high-speed video from a single compressive snapshot, event-stream representations of two time intervals (blue and red depict positive and negative spikes respectively), an image where the die appears stationary, as well as a motion-deblurred image.}
    \vspace{-0.05in}
    \label{fig:first-fig}
\end{figure*}

%% file: sections/related_work.tex
\paragraph{Reinterpretable imaging} has previously been explored at the granularity of light rays \cite{agrawal2010reinterpretable}, by modulating the plenoptic function, and at the level of spatio-temporal voxels~\cite{gupta2010flexible}, by using fast per-pixel shutters. 
SoDaCam represents a logical culmination of reinterpretability at the level of photon detections, that facilitates multiple post-capture imaging functionalities.

\paragraph{Programmable imaging} using a digital micromirror device was first introduced in \citet{nayar2006programmable} to perform \emph{pre-capture} radiometric manipulations. 
Modern programmable cameras are typically near-sensor processors \cite{scamp_2013,Wei_2018_ECCV,Tomohiro2017} that can perform limited operations in analog \cite{Martel2017,Wei_2018_ECCV,P2C2}, while more complex operations \cite{Martel2016,chen2017feature} occur after analog-to-digital conversion (ADC).
In contrast, by performing \emph{post-capture} computations directly on photon detections, we can perform complex operations without incurring the read-noise penalty that is associated with ADC.

\paragraph{Passive single-photon imaging.} Only recently have SPADs been utilized as passive imaging devices, with applications in high-dynamic range imaging \cite{ingleHighFluxPassive2019,inglePassiveInterPhotonImaging2021,Liu_2022_WACV,namiki2022imaging}, motion-compensation \cite{Iwabuchi2021,Seets_2021_WACV}, burst photography \cite{ma_quanta_2020,Ma_2023_WACV} and object tracking \cite{gyongy2018single}. Compared to compute-intensive burst-photography methods \cite{ma_quanta_2020}, our proposed techniques involve lightweight computations that can be performed near sensor.  
These computations can also be performed using other single-photon imagers such as Jots \cite{fossum2005sub,fossum2016quanta}, which feature higher sensor resolution and photon-efficiency \cite{Ma:17}, albeit at lower frame-rates and higher read noise.

\paragraph{Reducing the readout of SPADs.} Several data reduction strategies have been proposed in the context of SPADs that are used to timestamp incident photons, including: coarse histograms~\cite{della2020128,gyongy2020high,ren2018high}, compressive histograms~\cite{Gutierrez-Barragan_2022_CVPR}, and measuring differential time-of-arrivals~\cite{Zhang2022First,White2022DSPAD}. 
When SPADs are operated as photon-detectors, multi-bit counting \cite{morimoto_megapixel_2020}, or summing binary frames, can reduce readout.
While compression is not our main objective, we show that photon-cube projections act as camera-specific compression schemes that dramatically reduce sensor readout.

%% file: sections/background.tex
A SPAD array captures incident light as a \textit{photon-cube}: a temporal sequence of binary frames that represents the pixel-wise detection of photons across their respective exposure windows.
We can model the stochastic arrival of photons as a Poisson process \cite{yang_poisson_model}, allowing us to treat spatio-temporal values of the photon-cube as independent Bernoulli random variables with
\begin{equation}
\label{eq:shot_noise_model}
    \Prob{B_t(\vx) = 1} = 1 - e^{-\left(\eta \Phi(\vx, t)   + r_q\right)w_\text{exp}},  
\end{equation}
where $B_t(\vx)$ represents the value of the photon-cube at pixel $\vx$ and exposure index $1 \leq t \leq T$, which receives a mean incident flux of intensity $\Phi(\vx, t)$ across its exposure of duration $w_\text{exp}$.
Additionally, $\eta$ is the photon detection efficiency of the SPAD, and $r_q$ denotes the sensor's dark count rate---which is the rate of spurious counts unrelated to incident photons.
While individual binary frames are extremely noisy, the temporal sum of the {photon-cube}
\begin{equation}
\label{eq:sum_image}
    \cI_\text{sum}(\vx) := \sum_{t=1}^T B_t(\vx),
\end{equation}
can produce an `image' of the scene that is sharp in static regions, but blurry in dynamic regions (\cref{fig:exposure-stack} \textit{(top)}). Indeed, in static regions, the sum-image can be used to derive a maximum likelihood estimator of the scene intensity~\cite{Antolovic2016}, given by $\hat{\Phi}(\vx) = -\ln(1- T^{-1}{\cI_\text{sum}(\vx)})/{\eta w_\text{exp}} - {r_q}/{w_\text{exp}}$.

%% file: sections/projections.tex
The temporal sum described in \cref{eq:sum_image} is a simple instance of projections of a photon-cube. Our key observation is that it is possible to compute a wide range of photon-cube projections, each of which emulates a unique sensing modality \emph{post-capture}---including modalities that are difficult to achieve with conventional cameras. For example, varying the number of bit-planes that are summed over emulates exposure bracketing \cite{debevec2008recovering,mann1994beingundigital}, which is typically used for HDR imaging. Compared to conventional exposure bracketing, the emulated exposure stack, being software-defined, does not require spatial and temporal registration, which can often be error-prone. \cref{fig:exposure-stack} \textit{(top)} shows an example of an exposure stack computed from a photon-cube.

Going further, we can gradually increase the complexity of the projections. For example, consider a \emph{coded exposure} projection that multiplexes bit-planes with a temporal code
\begin{equation}
    \label{eq:flutter_shutter}
    \cI_\text{flutter}(\vx) := \sum_{t=1}^T C_t B_t(\vx),
\end{equation}
where $C_t$ is the temporal code. An example of globally-coded exposures is the flutter shutter camera \cite{raskar2006coded}, which uses pseudo-random binary codes for motion-deblurring.

More general coded exposures can be obtained via spatially-varying temporal coding patterns $C_t(\vx)$:
\begin{equation}
    \label{eq:video-comp}
    \cI_\text{coded}(\vx) := \sum_{t=1}^T C_t(\vx) B_t(\vx).
\end{equation}
\cref{fig:exposure-stack} \textit{(bottom)} shows spatially-varying exposures that use a quad (Bayer-like) spatial pattern and random binary masks. With photon-cubes, we can perform spatially-varying coding without bulky spatial light modulators, similar to focal-plane sensor-processors \cite{Martel2017,Wei_2018_ECCV}. Moreover, we can capture multiple coded exposures simultaneously, which is challenging to realize in existing sensors. In \cref{sec:video-compressive}, we describe coding patterns for video compressive sensing.
\input{figures/exposure_stack.tex}

Spatial and temporal gradients form the building blocks of several computer vision algorithms \cite{Canny1986,harris1988combined,horn1981determining,lucas1981iterative,dalal2005histograms}. 
Given this, another projection of interest is temporal contrast, i.e., a derivative filter preceded by a smoothing filter:
\begin{equation}
    \label{eq:temporal-deriv}
    \cI_\text{contrast}(\vx, t) := D_t \circ G * B_t(\vx),
\end{equation}
where $D_t$ is the difference operator, $G$ could be exponential or Gaussian smoothing, $\circ$ denotes function composition, and $*$ denotes convolution. 
Due to their sparse nature, gradients form the basis of bandwidth- and power-efficient event cameras \cite{lichtsteiner200364x64,serrano2013128,Finateau2020Prophesee,Chen_2019_CVPR_Workshops}, which we emulate in \cref{sec:event-camera}.

So far, we have considered projections taken only along the time axis. Next, we consider a more general class of \emph{spatio-temporal projections} that lead to novel functionalities. For instance, computing a simple projection, such as the temporal sum, along arbitrary spatio-temporal directions emulates sensor motion during exposure time~\cite{levin2008motion}, but \emph{without moving the sensor}. We achieve this by shifting bit-planes and computing their sum:
\begin{equation}
    \label{eq:motion_proj}
     \cI_\text{shift}(\vx):= \sum_{t=1}^T B_t\left(\vx + \vr(t) \right), 
\end{equation}
where $\vr$ is a discretized 2D trajectory that determines sensor motion. 
Outside a software-defined framework, such projections are hard to realize without physical actuators. 
We describe the capabilities of \textit{motion projections} in \cref{sec:motion-projection}.

In summary, the proposed photon-cube projections are simple linear and shift operators that lead to a diverse set of post-capture imaging functionalities. 
These projections pave the way for future `swiss-army-knife' imaging systems that achieve \emph{multiple functionalities (e.g., event cameras, high-speed cameras, conventional cameras, HDR cameras) simultaneously with a single sensor}. 
Finally, these projections can be computed efficiently in an online manner, which makes on-chip implementation viable (\cref{sec:ultraphase}). 

At this point, we note that a key enabling factor of photon-cube projections is the extremely high temporal-sampling rate of SPADs.
Indeed, the temporal sampling rate determines key aspects of sensor emulation, such as the discretization of temporal derivatives and motion trajectories.
This raises a natural question: can we use conventional high-speed cameras for computing projections? 

\paragraph{Trade-off between frame-rate and SNR.} In principle, photon-cube projections can be computed using regular (CMOS or CCD based) high-speed cameras. Unfortunately, each frame captured by a high-speed camera incurs a read-noise penalty, which increases with the camera's frame-rate~\cite{readout_noise}. In fact, the read noise levels of high-speed cameras \cite{phantom_v2640} can be $10$--$30\times$ higher than consumer cameras \cite{igual2019photographic}.  
Coupled with the low per-frame incident flux at high frame-rates, high levels of read noise result in extremely low SNRs.
In contrast, SPADs do not incur a per-frame read noise and are limited only by the fundamental photon noise. Hence, for the post-capture software-defined functionalities proposed here, it is imperative to use SPADs.

%% file: figures/exposure_stack.tex
\begin{figure}[t]
    \centering
    \includegraphics[width=\columnwidth]{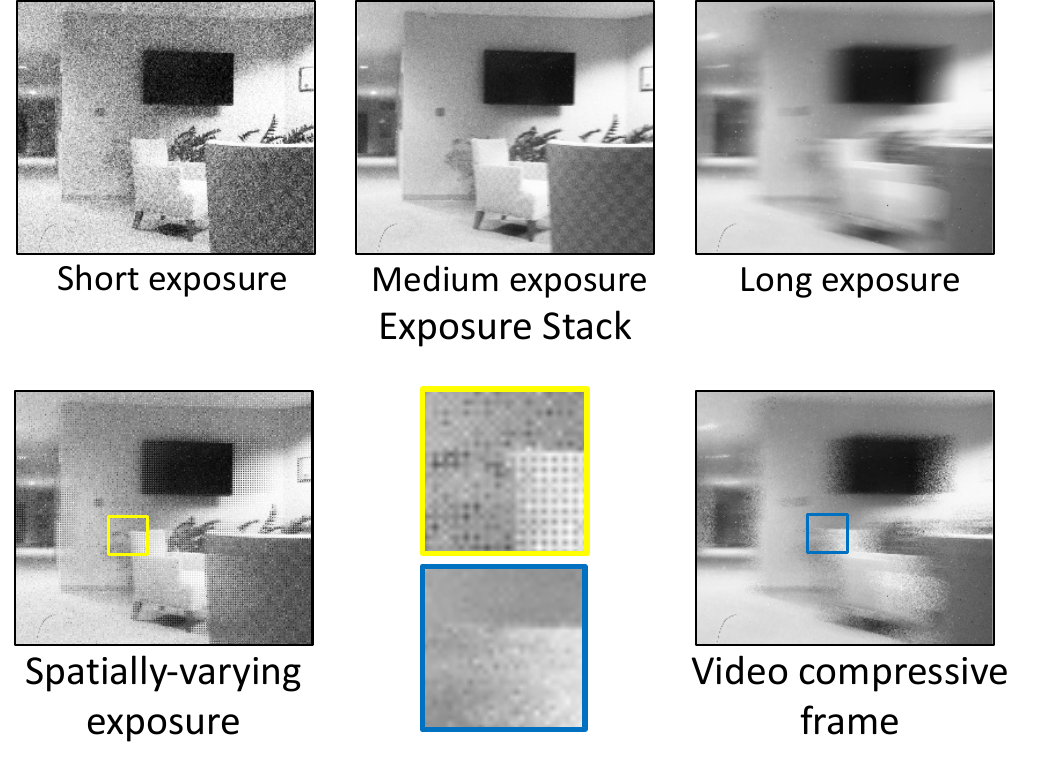}    
    \vspace{-0.15in}
    \tightcaption{\textbf{Coded exposures from photon-cubes.} \textit{(top)} An exposure stack with sum-images computed using $250$, $500$, and $1000$ bit-planes. Short exposures are noisy while long exposures exhibit motion blur. \textit{(bottom)} Spatially-varying exposure that uses a quad pattern \cite{jiang2021hdr} (see inset), and a video compressive frame that uses $16$ random binary masks to modulate the photon-cube. Zoom-in to see details.}
    \label{fig:exposure-stack}
\end{figure}

%% file: sections/emulating.tex
\cref{sec:projections} presented the concept of photon-cube projections and its potential for achieving multiple post-capture imaging functionalities. As case studies, we now demonstrate three imaging modalities: video compressive sensing, event cameras, and motion-projection cameras. These modalities have been well-studied over several years; in particular, there exist active research communities around video compressive sensing and event cameras today. We also show new variants of these imaging systems that arise from the software-defined nature of photon-cube projections. 

\subsection{Video Compressive Sensing}
\label{sec:video-compressive}
\input{sections/video_compressive.tex}

\subsection{Event Cameras}
\label{sec:event-camera}
\input{sections/event_cameras.tex}

\subsection{Motion Projections}
\label{sec:motion-projection}
\input{sections/motion_projections.tex}

%% file: sections/video_compressive.tex
Video compressive systems \emph{optically} multiplex light with random binary masks, such as the patterns in \cref{fig:vcs-masks} \textit{(left)}. As discussed in the previous section, such multiplexing can be achieved \emph{computationally} using photon-cubes.

\input{figures/vcs_mask.tex}

\paragraph{Two-bucket cameras.} One drawback of capturing coded measurements is the light loss due to blocking of incident light. To prevent loss of light, coded two-bucket cameras~\cite{Wei_2018_ECCV} capture an additional measurement that is modulated by the complementary mask sequence (\cref{fig:vcs-masks} \textit{(center)}). Such measurements recover higher quality frames, even after accounting for the extra readout~\cite{YuqiAndresonAcc2020,Shedligeri_2021_WACV}. Two-bucket captures can be readily derived from photon-cubes, by implementing \cref{eq:video-comp} with the additional mask sequence.

\paragraph{Multi-bucket cameras.} We can extend the idea of two-bucket captures to multi-bucket captures by accumulating bit-planes in one of $k$ buckets that is randomly chosen at each time instant and pixel location.  Since multiplexing is performed computationally, we do not face any loss in photoreceptive area that \cite{Woong4bucket2017,Wan2012MultiBucket} hampers existing multi-bucket sensors. Multi-bucket captures can reconstruct a large number of frames by better conditioning video recovery and provide extreme high-speed video imaging. \cref{fig:vcs-masks} \textit{(right)} shows the modulating masks for a four-bucket capture.

%% file: figures/vcs_mask.tex
\begin{figure}[t]
    \centering
    \includegraphics[width=\columnwidth]{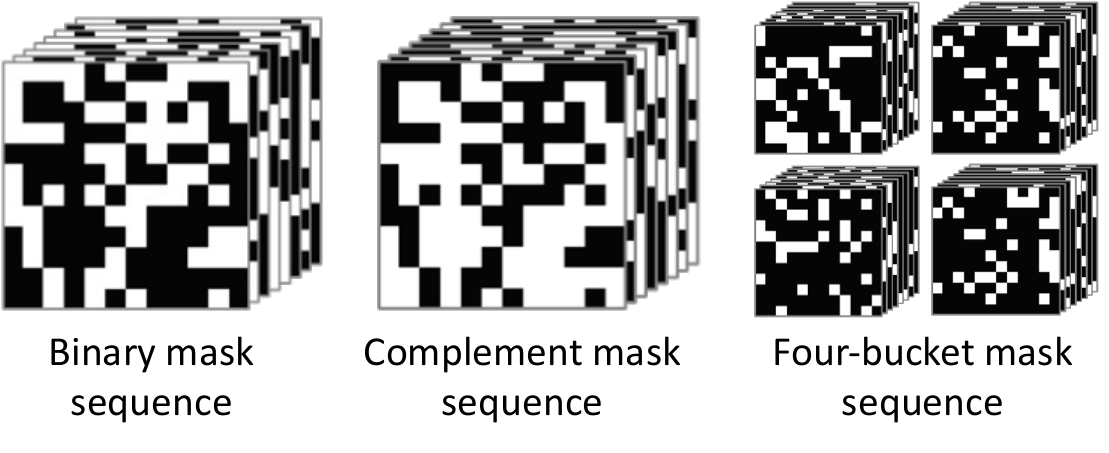}   
    \vspace{-0.15in}
    \tightcaption{\textbf{Modulating masks for video compressive sensing.} \textit{(left)} A single VCS measurement temporally compresses a sequence of frames using binary random masks. \textit{(center)} Two-bucket cameras capture an additional measurement by using the complementary mask sequence. \textit{(right)} We propose using multi-bucket captures by randomly choosing an active bucket for each frame. Both two-bucket and multi-bucket captures have $100$\% light efficiency. All masks are visualized here for $16 \times 16$ pixels.}
    \vspace{-0.05in}
    \label{fig:vcs-masks}
\end{figure}

%% file: sections/event_cameras.tex
Next, we describe the emulation of event cameras, which capture changes in light intensity and are conceptually similar to the temporal contrast projection introduced in \cref{eq:temporal-deriv}.
Physical implementations of event sensors~\cite{lichtsteiner200364x64,serrano2013128,Finateau2020Prophesee,Chen_2019_CVPR_Workshops} generate a photoreceptor voltage $V(\vx, t)$ with a logarithmic response to incident flux $\Phi(\vx, t)$, and output an event $(\vx, t, p)$ when this voltage deviates sufficiently from a reference voltage $V_\text{ref}(\vx)$:
\begin{equation}
\label{eq:event_principle}
    |V(\vx, t) - V_\text{ref}(\vx)| > \tau,
\end{equation}
where $\tau$ is called the contrast-threshold and $p=\text{sign}(V(\vx, t) - V_\text{ref}(\vx))$ encodes the polarity of the event. Once an event is generated, $V_\text{ref}(\vx)$ is updated to $V(\vx, t)$. \cref{eq:event_principle}, for a smoothly-varying flux intensity, thresholds a function of the temporal gradient, i.e., $\partial_t \log(\Phi(\vx, t))$.
\input{figures/event_cam.tex}

\paragraph{From bit-planes to event streams.} To produce events from SPAD frames, we compute an exponential moving average (EMA) of the bit-planes, as $\mu_t(\vx) = (1-\beta) B_t(\vx) + \beta \mu_{t-1}(\vx)$---where $\mu_t(\vx)$ is the EMA, $\beta$ is the smoothing factor, and $B_t$ is a bit-plane. We generate an event when $\mu_t(\vx)$ deviates from $\mu_\text{ref}(\vx)$ by at least $\tau$:
\begin{equation}
\label{eq:SPAD_event}
    |h(\mu_t(\vx)) - h(\mu_\text{ref}(\vx))| > \tau,
\end{equation}
where $h$ is a scalar function applied to the EMA. We can see that \cref{eq:SPAD_event} thresholds temporal contrast, by observing the role played by the EMA and the difference operator.

Setting $h$ to be the logarithm of the flux MLE mimics \cref{eq:event_principle}. However, since the log-scale is used to prevent sensor saturation, a simpler alternative is to use the non-saturating response curve of SPAD pixels ($h(x)=x$). The response curve takes the form of $1-\exp\left(-\alpha\Phi(\vx,t)\right)$, where $\alpha$ is a flux-independent constant. As a major advantage, this response curve avoids the underflow issues of the log function that can occur in low-light scenarios \cite{shi2022review}.

The SPAD's frame rate determines the time-stamp resolution of emulated events. In \cref{fig:event-cam}, we show the events generated from a photon-cube acquired at a frame-rate of $96.8$ kHz---resulting in a time-stamp resolution of $\tildeNice 10$ $\mu\text{s}$ that is comparable to those of existing event cameras.

\input{figures/motion_cam.tex}

\paragraph{How do SPAD-events differ from the output of a regular event camera?} The main difference is the expression of temporal contrast, given by $\partial_t h$, is now $-\partial_t \exp({-\alpha\Phi(\vx,t)})$, instead of $\partial_t \log(\Phi(\vx, t))$. 
This difference poses no compatibility issues for a large class of event-vision algorithms that utilize a grid of events~\cite{Rebecq_2019_CVPR,Scheerlinck_2020_WACV,Tulyakov_2021_CVPR} or brightness changes \cite{gehrig2020eklt}.  
We show examples of downstream applications using SPAD-events in \crefex{supl-sec:event-camera}.
Finally, SPAD-events can be easily augmented with spatially- and temporally-aligned intensity information---a synergistic combination that has been exploited by several recent event-vision works \cite{gehrig2020eklt,Hidalgo-Carrio_2022_CVPR,zhang2021object}.

%% file: figures/event_cam.tex
\begin{figure}[t]
    \centering
    \includegraphics[width=\columnwidth]{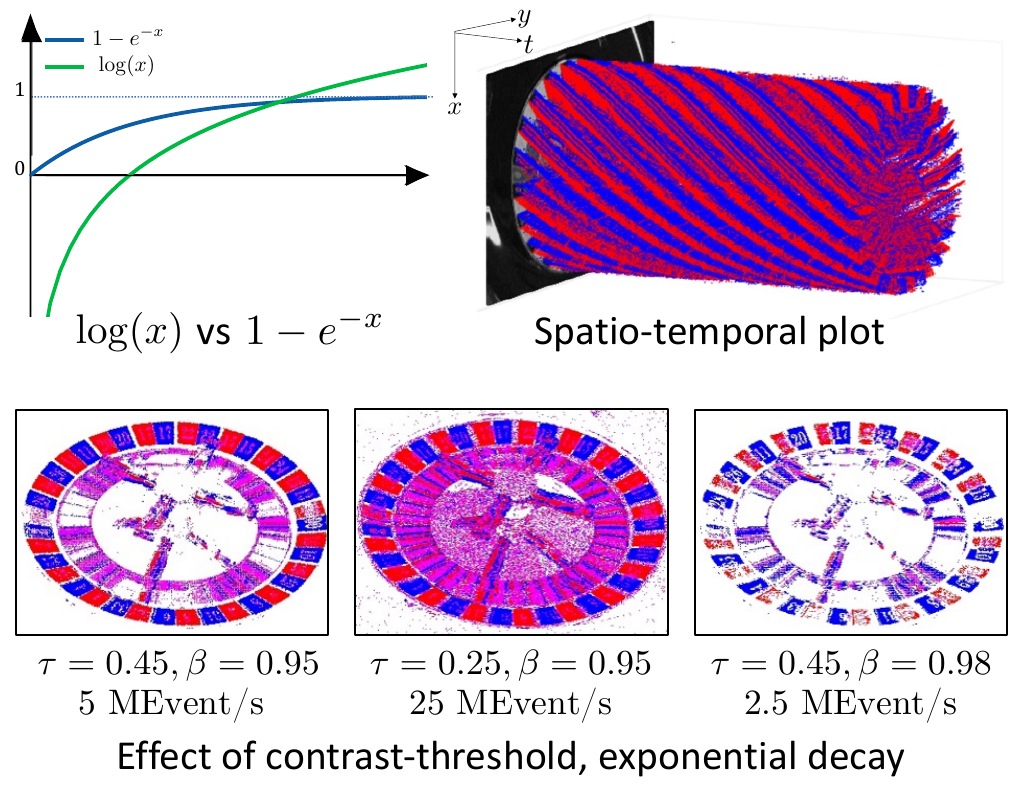}  
    \vspace{-0.15in}
    \tightcaption{\textbf{Event stream from photon-cubes.} \textit{(top left)} By exploiting the non-linear response curve of SPADs to encode brightness, we can avoid the underflow issues of a log-response. We visualize events generated from photon-cubes using a $3$D scatter plot of polarities (\textit{top right}, $14000$ bit-planes), and frame accumulation of events (\textit{bottom}, $1200$ bit-planes). Blue and red denote positive and negative spikes respectively. The event images also show the effect of varying the contrast threshold $\tau$ and exponential decay $\beta$---larger values yield a less noisy but sparser event stream.}
    \label{fig:event-cam}
\end{figure}

%% file: figures/motion_cam.tex
\begin{figure}[t]
    \centering
    \includegraphics[width=\columnwidth]{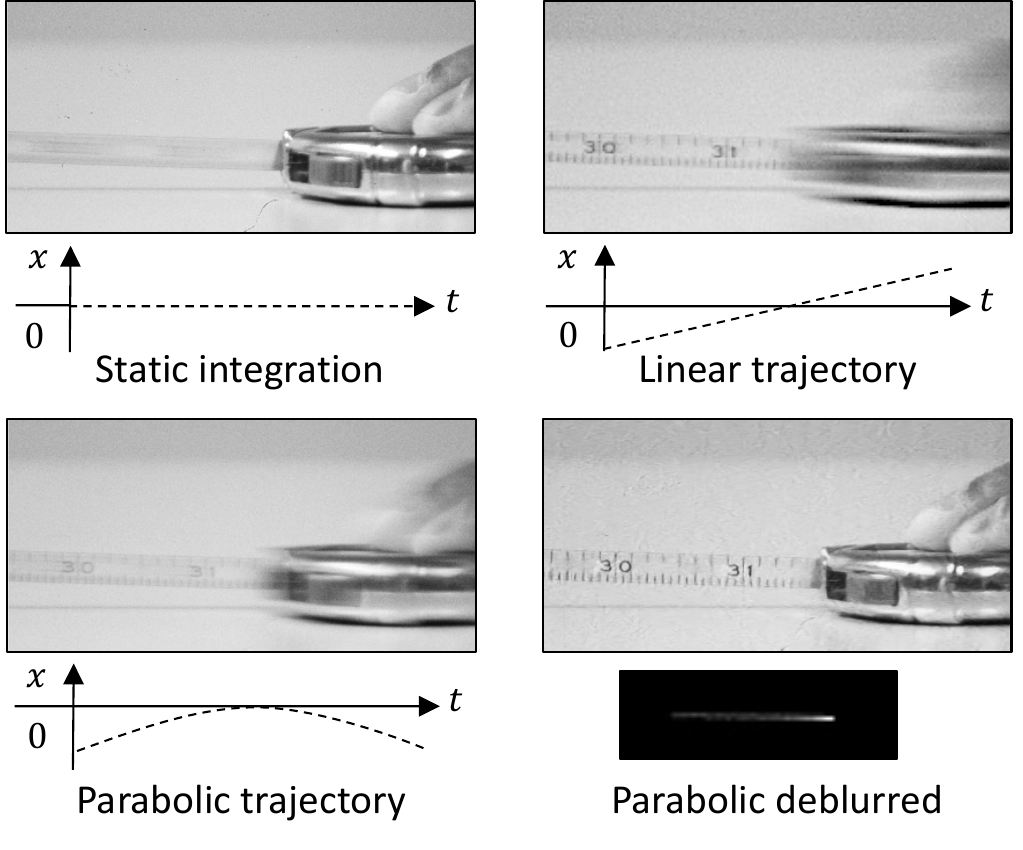}
    \vspace{-0.15in}
    \tightcaption{\textbf{Motion projections.} \textit{(top)} Integrating along a linear trajectory in the photon-cube changes the apparent image-space velocity of scene objects. Details are seen for \textit{(top left)} the case when static, and \textit{(top right)} the metallic tape when the sensor translates along the $x$-axis. \textit{(bottom)} A parabolic integration trajectory results in a motion-invariant image, resulting in similar blur kernels for all objects. \textit{(bottom right)} Deblurring with the resultant shift-invariant point spread function (shown in \textit{inset}) produces a sharp image.}
    \label{fig:motion-cam}
\end{figure}

%% file: sections/motion_projections.tex
Having described the emulation of cameras that capture coded exposures and temporal contrasts, we now shift our attention to cameras that emulate sensor motion during exposure, \textit{viz.} motion cameras. 
We describe two useful trajectories when emulating motion cameras using \cref{eq:motion_proj}.

\paragraph{Linear trajectory.}The simplest sensor trajectory involves linear motion, where $\vr(t) = (b t + c)\,\uvec{p}$ for some constants $b, c \in \rr$ and unit vector $\uvec{p}$. As \cref{fig:motion-cam} \textit{(top row)} shows, this can change the scene's frame of reference: making moving objects appear stationary and vice-versa.
\input{figures/motion_stack.tex}
\input{figures/compressive_video.tex}
\input{figures/motion_recons.tex}

\paragraph{Motion-invariant parabolic projection.} If motion is along $\uvec{p}$, parabolic integration produces a motion-invariant image~\cite{levin2008motion}---all objects, irrespective of their velocity, are blurred by the same point spread function (PSF), up to a linear shift. Thus, a deblurred parabolic capture produces a sharp image of all velocity groups (\cref{fig:motion-cam} \textit{(bottom row)}). The parabolic trajectory is given by $\vr(t) = (at^2 + b t + c)\,\uvec{p}$. We choose $a$ based on the maximum object velocity and $b, c$ so the parabola's vertex lies at $T/2$. We readily obtain the PSF by applying the parabolic integration to a delta input. Upon deconvolution using the PSF, a parabolic projection provides the optimal SNR for a blur-free image from single capture when only the direction of velocity is known. 

\paragraph{Ensembling linear projections.} Finally, we leverage the flexibility of photon-cubes to compute multiple linear projections, as seen in \cref{fig:motion-stack}. This produces a stack of images where one velocity group is motion-blur free at a time---or a `motion stack', analogous to a focal stack. This novel construct can be used to compensate motion by blending stack images using cues such as blur orientation or optical flow.

%% file: figures/motion_stack.tex
\begin{figure}[t]
    \centering
    \includegraphics[width=\columnwidth]{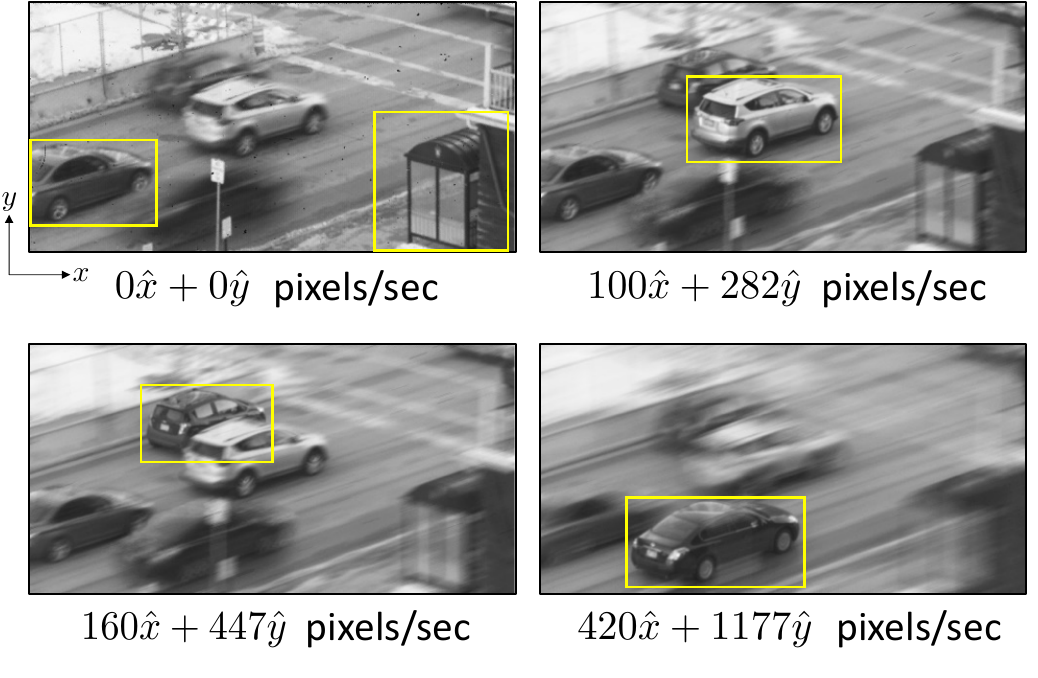} 
    \vspace{-0.2in}
    \tightcaption{\textbf{Motion stack.} Computing multiple linear projections with different trajectories can produce a stack of images where objects with matching velocities are sharp. 
    Here, we show a traffic scene involving four cars that have four different velocities.
    By suitably altering the slope of the linear trajectory, we can produce images where only one of the cars appears sharp at a time.
    We indicate the slope of the trajectories chosen and the objects that are ``in-focus''.}
    \label{fig:motion-stack}
\end{figure}

%% file: figures/compressive_video.tex
\begin{figure*}[t]
    \centering
    \includegraphics[width=\textwidth]{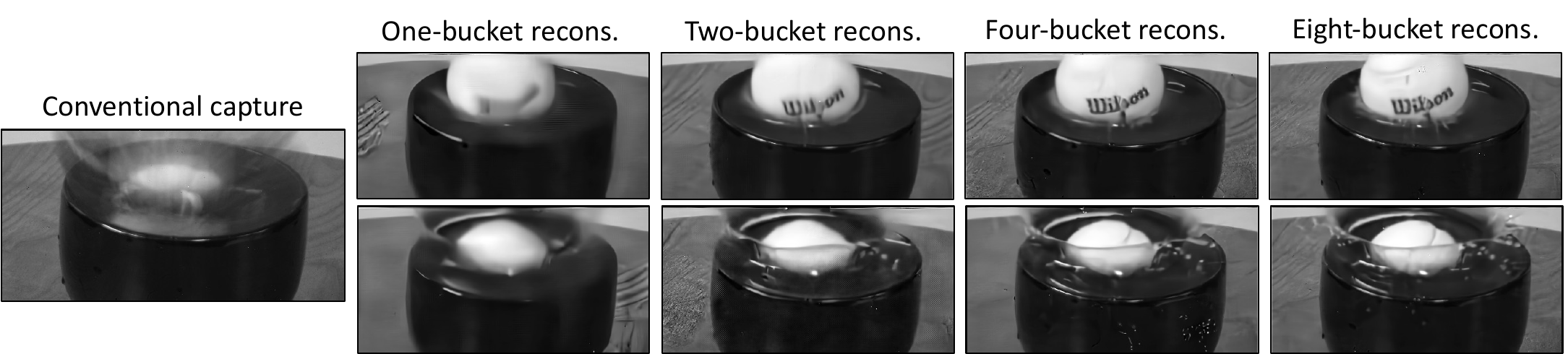} 
    \vspace{-0.1in}
    \tightcaption{\textbf{High-speed videography at $2000$ FPS} of a tennis ball dropped into a bowl of water, from a $25$ Hz readout. The conventional capture provides a visualization of the scene dynamics. It is challenging to reconstruct a large number of frames from a single compressive snapshot. Multi-bucket captures recover frames with significantly greater detail, such as the crown of water surrounding the ball. \textbf{We include more sequences (e.g., a bursting balloon) in the supplementary material.}}
    \vspace{-0.1in}
    \label{fig:comp-video}
\end{figure*}

%% file: figures/motion_recons.tex
\begin{figure*}[t]
    \centering
    \includegraphics[width=\textwidth]{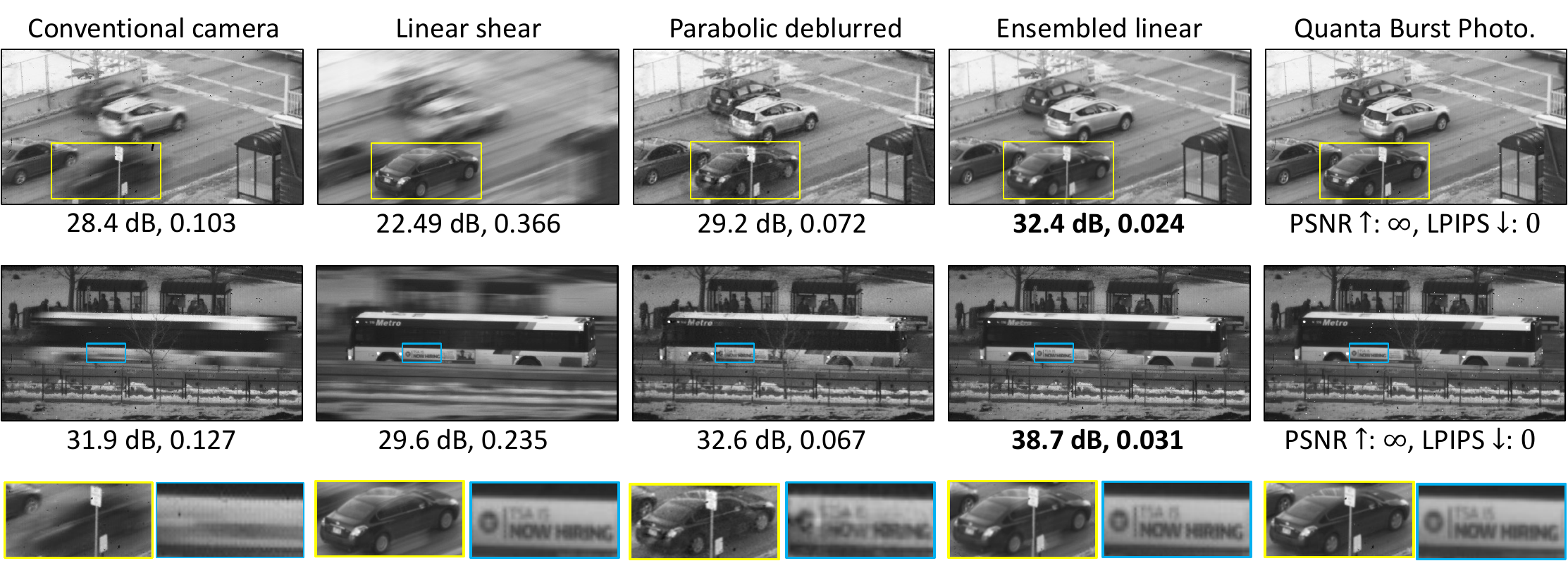}
    \vspace{-0.1in}
    \tightcaption{\textbf{Deblurring of traffic scenes using motion projections.} Linear projections can recover details of moving objects if their velocity is known. When only the motion direction is known (e.g., road's orientation), a sharp image can be obtained by either deblurring a parabolic projection or by blending multiple randomly-sampled linear projections. We quantitatively compare against the compute- and bandwidth-expensive Quanta Burst Photography \cite{ma_quanta_2020}, based on PSNR and LPIPS \cite{Zhang_2018_CVPR}.}
    \label{fig:motion-recons}
\end{figure*}

%% file: sections/experiments.tex
We design a range of experiments to demonstrate the versatility of photon-cube projections: both when computations occur after readout (\cref{sec:comp-real-hardware,sec:swiss-spad2}), and when they are performed near-sensor on-chip (\cref{sec:ultraphase}). All photon-cubes were acquired using the  SwissSPAD2 array~\cite{ulku512512SPAD2019}, operated using one of two sub-arrays, each having $512 \times 256$ pixels, and at a frame-rate of $96.8$ kHz. For the on-chip experiments, we use the UltraPhase compute architecture to interface with photon-cubes acquired by the SwissSPAD2. 

\subsection{SoDaCam Capabilities}
\label{sec:swiss-spad2}

\paragraph{High-speed compressive imaging.} We reconstruct $80$ frames from compressive snapshots that are emulated at $25$ Hz, resulting in a $2000$ FPS video.  We decode compressive snapshots using a plug-and-play (PnP) approach, PnP-FastDVDNet \cite{Yuan2022PnPADMM}. As \cref{fig:comp-video} shows, it is challenging to recover a large number of frames from a single compressive measurement. Using the proposed multi-bucket scheme significantly improves the quality of video reconstruction. While multi-bucket captures require more bandwidth, this can be partially amortized by coding only dynamic regions, which we show in \crefex{supl-sec:video-comp}.

\paragraph{Motion projections on a traffic scene.} \cref{fig:motion-recons} shows two traffic scenes captured using a $50$ mm focal length lens and at $30$ Hz emulation. When object velocity is known, a linear projection can make moving objects appear stationary. If only the velocity direction is known (e.g., road's orientation in \cref{fig:motion-recons}), a parabolic projection provides a sharp reconstruction of all objects. We deblur parabolic captures using PnP-DnCNN \cite{zhang2020plug}. We offer an improvement by randomly sampling $8$ linear projections along the velocity direction and blending them using the optical flow predicted by RAFT \cite{teed2020raft} between two short exposures.
\input{figures/prophesse_comp.tex}

\paragraph{Low-light event imaging.} \cref{fig:prophesse-comp} compares event-image visualizations of SPAD and that of a state-of-the-art commercial event sensor (Prophesee EVK4), across various light levels, with an accumulation period of $33$ ms. For a fair comparison, we bin the Prophesee's events in blocks of $2 \times 2$ pixels and use a smaller aperture to account for the lower fill factor of the SPAD. We tuned event-generation parameters (contrast threshold, integrator decay rate) of both cameras at each light level. Low light induces blur and deteriorates the Prophesee's event stream. In contrast, SPAD-events continue to capture temporal gradients, due to the SPAD's low-light capabilities and its brightness-encoding response curve. We include an ablative study of brightness-encoding functions in \crefex{supl-sec:event-camera}.

Our observations are in concurrence with recent works that examine the low-light performance of event cameras~\cite{Hu_2021_CVPR,graca2021unraveling}, and show that SPAD-events can provide neuromorphic vision in these challenging-SNR scenarios.

\subsection{Comparison to High-Speed Cameras}
\label{sec:comp-real-hardware}

\input{figures/infinicam_comp.tex}

Recall, as previously discussed in \cref{sec:projections}, that read-noise limits the per-frame SNR of high-speed cameras. To demonstrate this limitation, we compute projections using the $4$ kHz acquisition of the Photron Infinicam, a conventional high-speed camera, at a resolution of $1246 \times 240$ pixels. We operate the SwissSPAD2 and the Infinicam at ambient light conditions using the same lens specifications. As \cref{fig:infinicam-comp} shows, read noise corrupts the incident signal in Infinicam and makes it impossible to derive any useful projections. The read noise could be averaged out to some extent if the Infinicam did not perform compression-on-the-fly, but compression is central to the camera's working and enables readout over USB. Using a larger aperture to admit more light improves the quality of computed projections, but the video reconstruction and event image remain considerably worse than the corresponding outputs of the SPAD.

\subsection{Bandwidth and Power Implications}
\label{sec:ultraphase}

\input{figures/ultraphase.tex}

While \cref{sec:swiss-spad2} has demonstrated the capabilities of photon-cube projections, we now show that our projections can also be obtained in a bandwidth-efficient manner via near-sensor computations. 
We implement photon-cube projections on UltraPhase (\cref{fig:ultraphase} \textit{(left)}), a novel compute architecture designed for single-photon imaging. 
UltraPhase consists of $3 \times 6$ processing cores, each of which interfaces with $4\times 4$ pixels, and can be 3D stacked beneath a SPAD array. 
We include visualizations and programming details of a few example projections in \crefex{supl-sec:ultraphase}.

We measure the readout and power consumption of UltraPhase when computing projections on $2500$ bit-planes of the falling die sequence (\cref{fig:first-fig}). 
The projections include: VCS with $16$ random binary masks, an event camera, a linear projection, and a combination of the three. 
We output projections at $12$-bit depth and calculate metrics based on the clock cycles required for both compute and readout. 
As seen in \cref{fig:ultraphase} \textit{(right)}, computing projections on-chip dramatically reduces sensor readout and power consumption as compared to reading out the photon-cube. 
Finally, similar to existing event cameras, SPAD-events have a resource footprint that reflects the underlying scene dynamics.

In summary, our on-chip experiments show that performing computations near-sensor can increase the viability of single-photon imaging in resource-constrained settings.

%% file: figures/prophesse_comp.tex
\begin{figure}[t]
    \centering
    \includegraphics[width=\columnwidth]{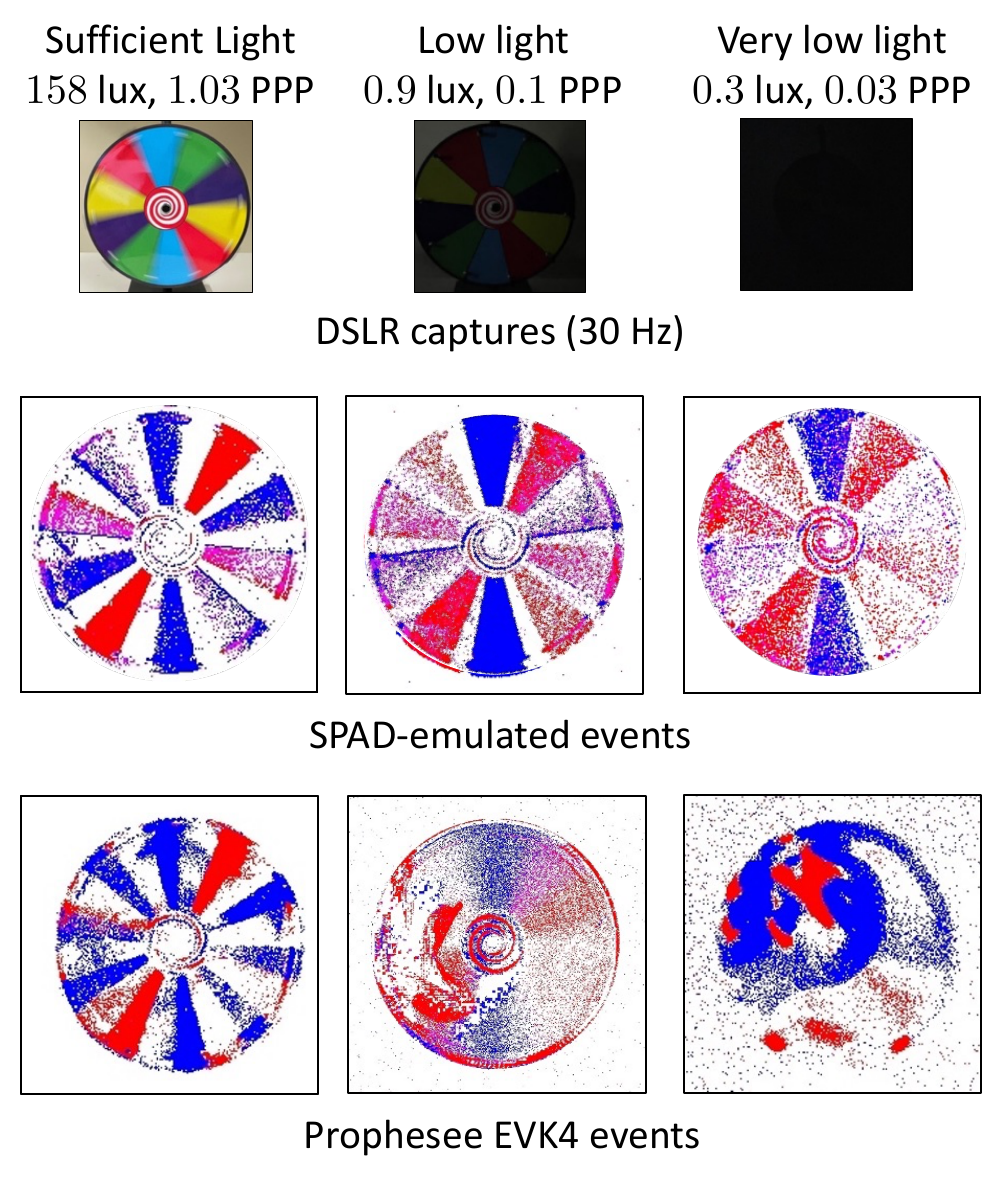}    
    \vspace{-0.2in}
    \tightcaption{\textbf{Comparison to a state-of-the-art event camera.} SPAD-events can capture temporal gradients even when the light-level is reduced by $500\times$, by benefiting from their single-photon sensitivity and bounded brightness response curve. In contrast, low-light induces blur and deteriorates the Prophesee's event stream. As a measure of the light-level, we report the PPP (photons per pixel) averaged across bit-planes and a light meter's reading at the sensor location.}
    \label{fig:prophesse-comp}
\end{figure}

%% file: figures/infinicam_comp.tex
\begin{figure}[t]
    \centering
    \includegraphics[width=\columnwidth]{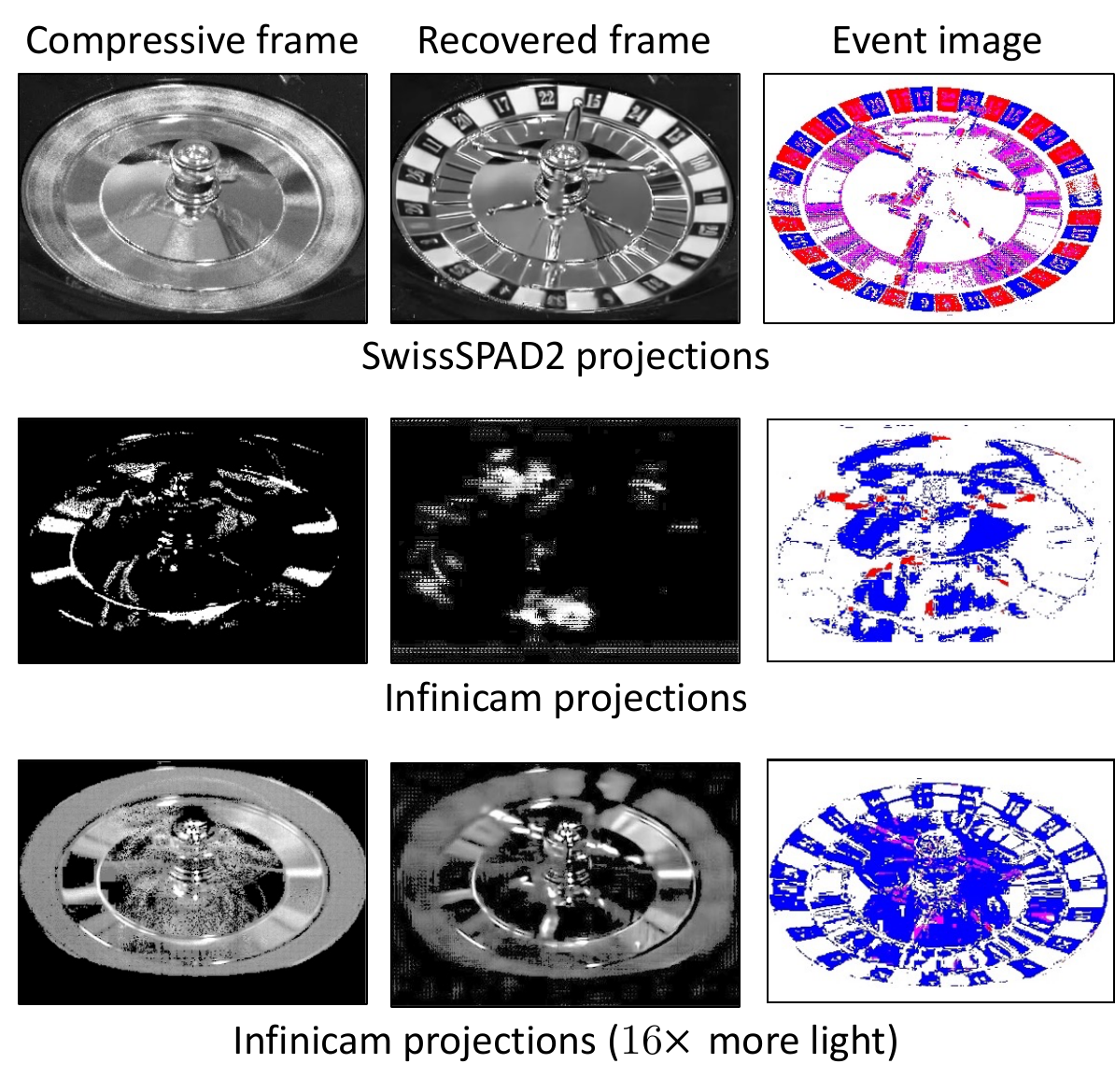}   
    \vspace{-0.2in}
   \tightcaption{\textbf{Comparison against conventional high-speed acquisition at $4000$ Hz.} \textit{(top)} SPAD projections recover a $16\times$ compressive video and an event image of a spinning roulette wheel. \textit{(middle)} Read-noise corrupts the incident flux in the Infinicam high-speed camera, removing details in frames which are compressed on-the-fly. \textit{(bottom)} Although using a larger aperture to admit more light recovers some detail, noise and compression artifacts still persist.}
    \label{fig:infinicam-comp}
\end{figure}

%% file: figures/ultraphase.tex
\begin{figure}[t]
    \centering
    \includegraphics[width=\columnwidth]{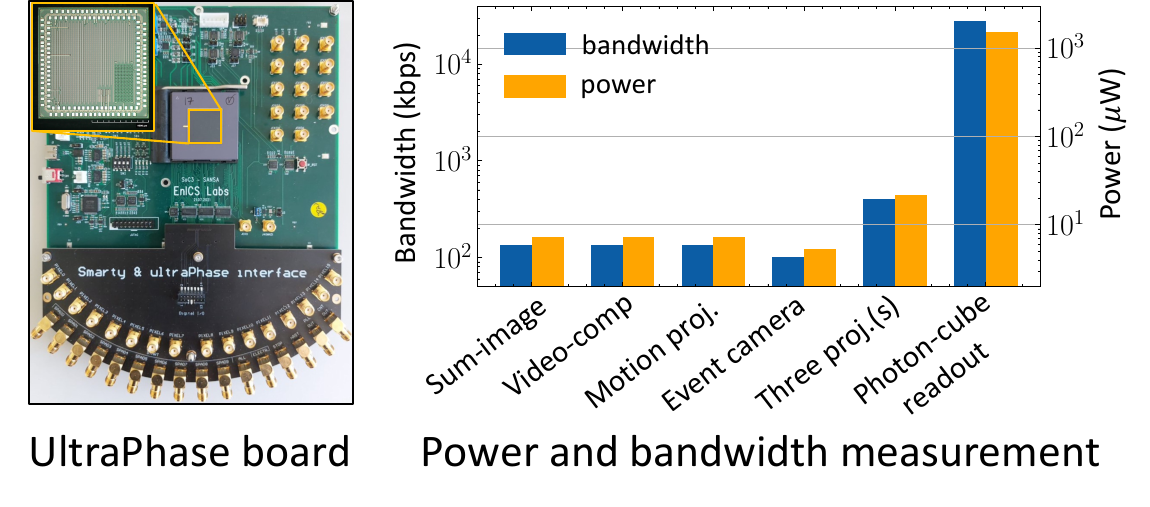} 
    \vspace{-0.15in}
    \tightcaption{\textbf{Power and bandwidth requirements} when computing photon-cube projections on UltraPhase \cite{ardelean2023computational} \textit{(left)}, a recent compute architecture designed for single-photon imaging, at $40$ Hz readout. \textit{(right)} Our projections act as a compression scheme for photon-cubes, resulting in dramatically reduced sensor readout and power consumption.}
    \vspace{-0.05in}
    \label{fig:ultraphase}
\end{figure}

%% file: sections/discussion.tex
SoDaCam provides a realization of reinterpretable software-defined cameras \cite{nayar2006programmable,gupta2010flexible,agrawal2010reinterpretable,adelson1991plenoptic,levoy1996light} at the fine temporal resolution of SPAD-acquired photon-cubes. The proposed computations, or photon-cube projections, can match and in some cases, surpass the capabilities of existing imaging systems. The software-defined nature of photon-cube projections provides functionalities that may be difficult to achieve in conventional sensors. These projections can reduce the readout and power-consumption of SPAD arrays and potentially spur widespread adoption of single-photon imaging in the consumer domain. Finally, future chip-to-chip communication standards may also make it feasible to compute projections on a camera image signal processor.

\paragraph{Adding color to SoDaCam.} One way to add color is by overlaying color filter arrays (CFAs) and perform demosaicing on the computed photon-cube projection: depending on the projection, demosaicing could be relatively simple or more complex. 
As a reference, Bayer CFAs have been considered in the context of both video compressive sensing \cite{Yuan2022PnPADMM} and event cameras \cite{GemmacolorDAVIS}. Incorporating CFAs with motion projections requires careful considerations, e.g., avoiding integrating across pixel locations of differing color. 

\paragraph{Future outlook on SPAD characteristics.} A key SPAD characteristic that determines several properties of emulated cameras is the frame rate. 
While no fundamental limitations prevent SPADs from being operated at the frame rates utilized in this work ($\tildeNice 100$ kHz), sensor readout and power constraints can preclude high speeds, especially in high-resolution SPAD arrays.
Photon-cube projections can enable future large-format SPADs to preserve high-speed information with modest resource requirements.

\paragraph{A platform for comparing cameras.} Comparing imaging modalities can be quite challenging since hardware realizations of sensors can differ in numerous aspects, such as their quantum efficiency, fill factor, pixel pitch, and array resolution.
By emulating their imaging models, SoDaCam can serve as a platform for hardware-agnostic comparisons; for instance, determining operating conditions where one imaging modality is advantageous over another.

\paragraph{A Cambrian explosion of new cameras.} Besides comparing cameras, by virtue of being software-defined, SoDaCam can also make it significantly easier to prototype and deploy new unconventional imaging models, and even facilitate sensor-in-the-loop optimization~\cite{sitzmann2018end,Metzler_2020_CVPR,martel2020neural} by tailoring photon-cube projections for downstream computer-vision tasks. This is an exciting future line of research.

%% file: supplementary.tex
\section{Video Compressive Sensing}\label{supl-sec:video-comp}
\input{supplementary_sections/video_compressive.tex}

\clearpage

\section{Event Cameras}\label{supl-sec:event-camera}
\input{supplementary_sections/event_cameras.tex}
\clearpage

\section{Motion Projections}\label{supl-sec:motion-proj}
\input{supplementary_sections/motion_projection.tex}
\clearpage

\section{Experimental Setup for \cref{sec:swiss-spad2,sec:comp-real-hardware}}\label{supl-sec:setup}
\input{supplementary_sections/experimental_setup.tex}
\clearpage

\section{UltraPhase Experiments}\label{supl-sec:ultraphase}
\input{supplementary_sections/ultraphase.tex}
\clearpage

\renewcommand\refname{Supplementary References}
{\small

\input{supp.bbl}
}

%% file: supplementary_sections/video_compressive.tex
In this supplementary note, we provide the pseudo code that describes the emulation of two- and multi-bucket cameras and mathematically describe their multiplexing masks. We also specify algorithmic details for video recovery from compressive measurements.

\subsection*{Multi-Bucket Capture Pseudocode}

\Cref{alg:multi-bucket} describes the emulation of $J$-bucket captures, denoted as $\cI_\text{coded}^j(\vx)$, from the photon-cube $B_t(\vx)$ using multiplexing codes $C^j_t(\vx)$, where $1\leq j \leq J$. Both single compressive snapshots (or one-bucket captures) and two-bucket captures can be emulated as special cases of \cref{alg:multi-bucket}, with $J=1$ and $J=2$ respectively.

\begin{algorithm}
\caption{Multi-Bucket Capture Emulation}\label{alg:multi-bucket}
\begin{algorithmic}
\Require {Photon-cube $B_t(\vx)$ \\
Number of buckets $J$\\
Multiplexing code for $j$\textsuperscript{th} bucket, $1 \leq j \leq J$, $C_t^j(\vx)$\\
Pixel locations $\mathcal{X}$\\
Total bit-planes $T$}
\Ensure {Multiplexed captures $\cI_\text{coded}^j(\vx)$}
\Function{MultiBucketEmulation}{$B_t(\vx)$, $C_t^j(\vx)$}
\State $Y^j(\vx) \gets 0, \,\forall\, j$
\For{$\vx \in \cx$, $1\ \leq j \leq J$}
\For {$1 \leq t \leq T$}
\State $\cI_\text{coded}^j(\vx) \gets \cI_\text{coded}^j(\vx) + B_t(\vx) \cdot C_t^j(\vx)$
\EndFor
\EndFor\\
\Return $\cI_\text{coded}^j(\vx)$
\EndFunction
\end{algorithmic}
\end{algorithm}

\paragraph{Mask sequences for video compressive sensing.} For a single compressive capture ($J=1$), a sequence of binary random is used, i.e, $C^1_t(\vx) = 1$ with probability $0.5$. For a two bucket capture, we use $$C^2_t(\vx) = 1 - C^1_t(\vx),$$which is the complementary mask sequence. For $J > 2$, at each timestep $t$ and pixel location $\vx$, the active bucket is chosen at random:$$C^j_t(\vx) \gets 1,\; j \sim \text{Uniform}(1, J).$$ This is a direct generalization of the masking used for both one- and two-bucket captures.

\subsection*{Decoding Video Compressive Captures}

A variety of decoding algorithms have been developed for video compressive sensing, including: (a) optimization frameworks tailored to the forward model of \cref{eq:video-comp} and with additional regularization  \cite{Liu19DeSCI,gapTVYuan2016}, (b) end-to-end deep-learning methods that utilize a large corpus of training data \cite{Yuan2022ELPUnfolding,Shedligeri_2021_WACV,Wang_2021_CVPR,Wu_2021_ICCV,YuqiAndresonAcc2020}, and (c) hybrid, plug-and-play (PnP) approaches that utilize an optimization framework but perform one or more steps using a deep denoiser \cite{chan2017PnP,venkatakrishnan2013PnP,Yuan2022PnPADMM}. We opt to use the PnP approach featuring an ADMM formulation \cite{yuan2021snapshot} and a deep video denoiser (FastDVDNet \cite{Tassano_2020_CVPR}) in this work. We justify our choice by noting that PnP-ADMM can produce high-quality reconstructions, comparable to end-to-end counterparts while using an off-the-shelf denoiser---precluding the need to train separate models for various masking strategies. 

For computational efficiency, PnP-ADMM requires the gram matrix of the resulting linear forward model of \cref{eq:video-comp} to be efficiently invertible. The multi-bucket scheme described above adheres to this consideration.

\subsection*{Constant-Bandwidth Comparison}

We now present a comparison of single-, two- and multi-bucket compressive captures when the readout rate is fixed. As Supp. Fig. \ref{supl-fig:constant-bandwidth} shows, multi-bucket captures provide higher fidelity reconstructions even when bandwidth is fixed. Furthermore, their bandwidth cost can be amortized, to some extent, by coding only dynamic regions---we describe this next.

\input{supplementary_figures/constant_bandwidth.tex}

\subsection*{Coding Only Dynamic Regions: Mitigating the Bandwidth Cost of Multi-Bucket Captures}

We observe that multi-bucket captures capture redundant information in static regions of the scene since each pixel in a static region has the same expected value under random binary modulation. Hence, we propose coding only dynamic regions---the dynamic region-of-interest (RoI) can be determined by masking pixels whose coded exposures deviate significantly from one another. As seen in Supp. Fig. \ref{supl-fig:code-dynamic}, the dynamic content may just be $25$\% of the image area, which provides significant scope for bandwidth savings. We observe that we can code just $25$\% of the total pixels, among additional compressive measurements, without a perceptible drop in visual quality, which yields an overall bandwidth requirement of $1.45\times$, or under twice the bandwidth cost of a single compressive measurement.

\input{supplementary_figures/code_dynamic.tex}

\subsection*{Results on More Sequences}

Additional results are shown in Supp. Fig. \ref{supl-fig:video-more}. 

\input{supplementary_figures/video_recons_more.tex}

%% file: supplementary_figures/constant_bandwidth.tex
\begin{figure*}[htp]
    \centering
    \includegraphics[width=\textwidth]{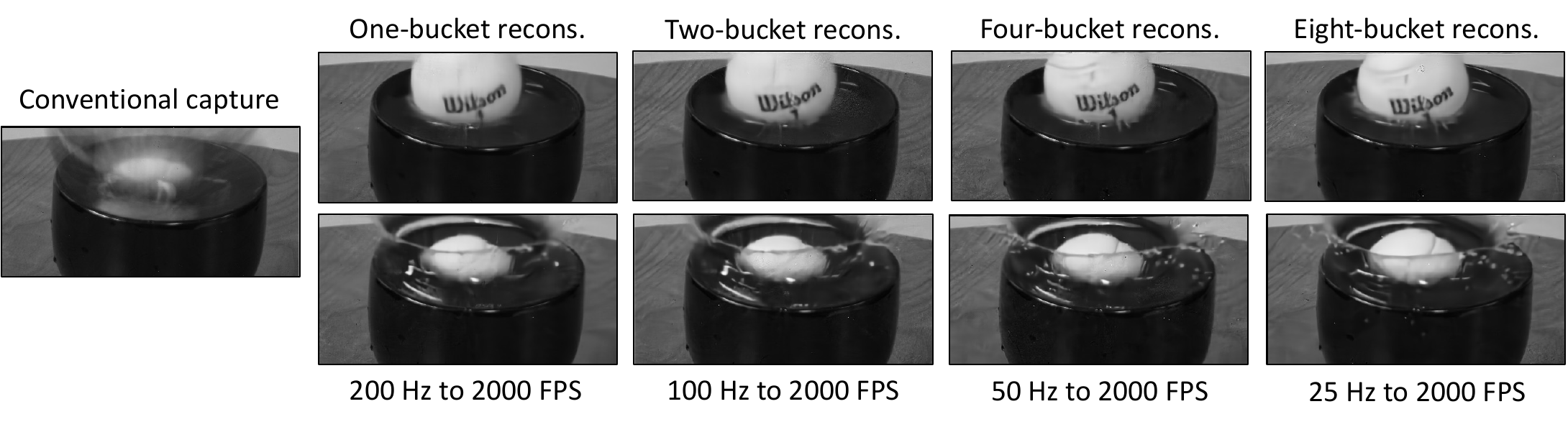} 
    \vspace{-0.15in}
    \tightcaption{\textbf{Fixed readout comparison of compressive video schemes.} We compare video reconstruction obtained from a single compressive snapshot, two-bucket capture, four-bucket capture and eight-bucket capture while holding readout constant---we achieve this by commesurately increasing readout, for instance, by reading out single compressive snapshots at $200$ Hz. We indicate the readout rate here in Hertz (Hz) and the frame-rate of the reconstructed video in FPS. Clearly, multi-bucket captures provide better reconstruction results than a burst of independently multiplexed captures.}
    \label{supl-fig:constant-bandwidth}
\end{figure*}

%% file: supplementary_figures/code_dynamic.tex
\begin{figure*}[htp]
    \centering
    \includegraphics[width=\textwidth]{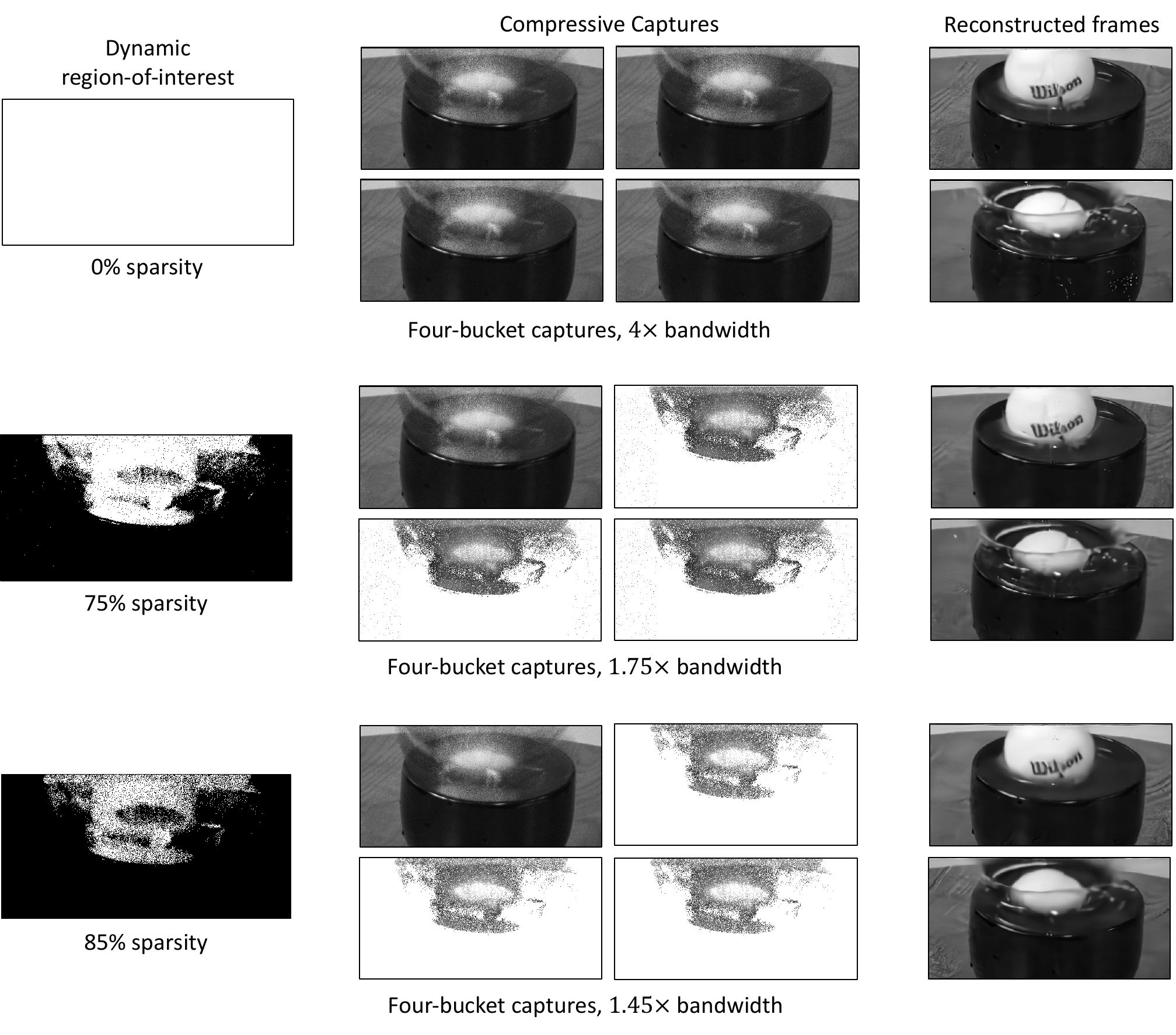} 
    \vspace{-0.15in}
    \tightcaption{\textbf{Coding dynamic regions can reduce readout of multi-bucket captures.} \textit{(left column)} Dynamic regions are detected by computing the standard deviation of the coded exposures and thresholding them appropriately (e.g., by the $75^\text{th}$ percentile)---we show the dynamic regions in white here. Coded exposures are transmitted only in the dynamic regions. For the static regions, we simply use a long exposure---by adding multi-bucket captures. \textit{(right column)} We observe that readout-bandwidth can be reduced to $1.75 \times$ from $4\times$ in the case of a four-bucket capture with no perceptual degradation of reconstruction quality. Bandwidth is provided here as a multiple of the readout of a single compressive capture.}
    \label{supl-fig:code-dynamic}
\end{figure*}

%% file: supplementary_figures/video_recons_more.tex
\begin{figure*}[htp]
    \centering
    \includegraphics[width=\textwidth]{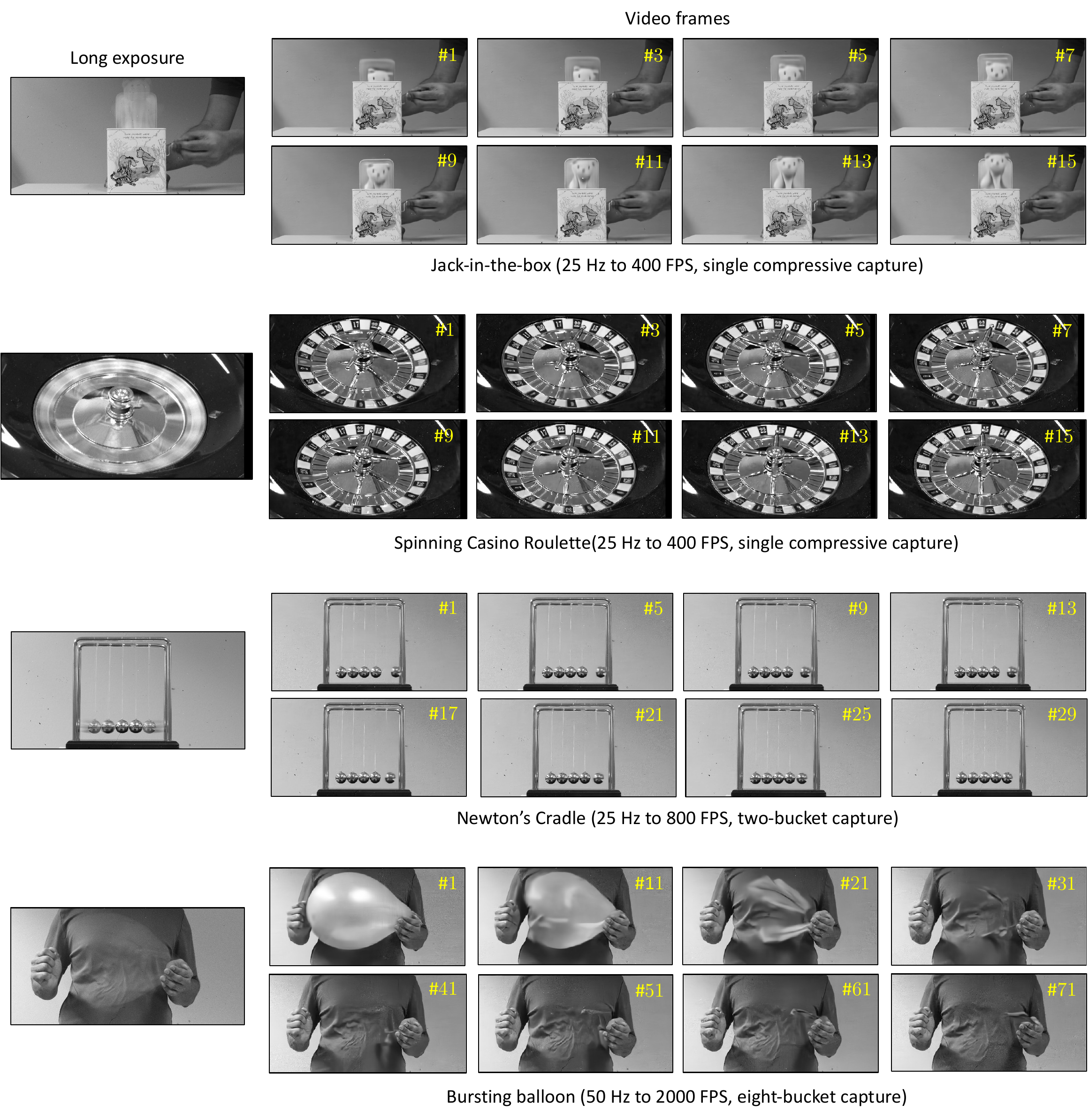} 
    \vspace{-0.15in}
    \tightcaption{\textbf{Results on additional sequences.} We use Hertz (Hz) to indicate the rate of emulation and frames-per-second (FPS) to indicate the frame-rate of the reconstructed video. Frame numbers are indicated in yellow font.}
    \label{supl-fig:video-more}
\end{figure*}

%% file: supplementary_sections/event_cameras.tex
\subsection*{Event-Generation Pseudocode}

We provide the pseudocode for emulating events from photon-cubes in \cref{alg:event-camera}. The contrast threshold $\tau$ and exponential smoothing factor $\beta$ are the two parameters that determine the characteristics of the resulting event stream, such as its event rate (number of events per second). We use an initial time-interval $T_0$ (typically $80$--$100$ bit-planes) to initialize the reference moving average, with $T_0$ being much smaller than $T$. The result of this pseudocode is an event-cube, $E_t(\vx)$, which is a sparse spatio-temporal grid of event polarities---positive spikes are denoted by $1$ and negative spikes by $-1$. From the emulated event-cube, other event representations can be computed such as: an event stream, $\{(\vx, t, p)\}$, where $p \in \{-1, 1\}$ indicates the polarity of the event; a frame of accumulated events~\cite{Maqueda_2018_CVPR} (seen in \cref{fig:event-cam,fig:prophesse-comp}); and a voxel grid representation~\cite{zhu2019unsupervised}, where events are binned into a few temporal bins (shown in Supp. Fig. \ref{supl-fig:event-algo} \textit{(top left)} using $3$ temporal bins).

\begin{algorithm}
\caption{Event Camera Emulation}\label{alg:event-camera}
\begin{algorithmic}
\Require {Photon-cube $B_t(\vx)$ \\
Contrast threshold $\tau$\\
Exponential smoothing factor, $\beta$\\
Pixel locations $\mathcal{X}$\\
Initial time-interval $T_0$, for computing reference moving average\\
Total bit-planes $T$}
\Ensure {Event-cube $E_t(\vx)$ that describes the spatio-temporal spikes}
\Function{EventCameraEmulation}{$B_t(\vx)$, $\tau$, $\beta$, $T_0$}
\State $E_t(\vx) \gets 0, \,\forall\, t, \, \forall \, \vx$
\For{$\vx \in \cx$}
\State Reference moving average, $\mu_\text{ref}(\vx) \gets 0$
\State Current moving average, $\mu_0(\vx) \gets 0$
\For {$1 \leq t \leq T_0$}
\State $\mu_\text{ref}(\vx) \gets \beta \mu_\text{ref}(\vx) + (1-\beta) B_t(\vx)$
\EndFor
\For {$T_0 \leq t \leq T$}
\State $\mu_t(\vx) \gets \beta \mu_{t-1}(\vx) + (1-\beta) B_t(\vx)$
\If{$\abs{\mu_t(\vx) - \mu_\text{ref}(\vx)} > \tau$}
\State $E_t(\vx) \gets \text{sign}(\mu_t(\vx) - \mu_\text{ref}(\vx))$
\State $\mu_\text{ref}(\vx) \gets \mu_\text{ref}(\vx) + \tau* \text{sign}(\mu_t(\vx) - \mu_\text{ref}(\vx))$
\EndIf
\EndFor
\EndFor\\
\Return $E_t(\vx)$
\EndFunction
\end{algorithmic}
\end{algorithm}

\subsection*{Compatibility of SPAD-Events with Existing Event-Vision Algorithms}

We now provide examples of downstream algorithms applied to SPAD-events, which shows the compatibility of the emulated event streams with existing event-vision algorithms.
Supplementary Figure \ref{supl-fig:event-algo} shows three downstream algorithms with SPAD-events as their input: Contrast Maximization~\cite{Gallego_2018_CVPR} which generates a warped image of events that has sharp edges (\textit{top right}), E2VID~\cite{Rebecq_2019_CVPR} which estimates intensity frames from an event stream (\textit{bottom left}), and DCEIFlow~\cite{DCEIFlow2022} which computes dense optical flow using intensity frames and aligned events (\textit{bottom right}). Both E2VID and DCEIFlow use a voxel grid representation of events as their inputs. We include the visualization of a voxel grid representation in Supp. Fig. \ref{supl-fig:event-algo} \textit{(top left)}.
All event streams were emulated using $3000$ bit-planes of photon-cubes acquired at $96.8$ kHz, and using $\beta=0.95$ and $\tau=0.4$ as emulation parameters.
We note that the performance of these algorithms can be improved by finetuning pre-trained learning-based models on a dataset of SPAD-events.

\input{supplementary_figures/event_algo.tex}

\subsection*{Ablation of Brightness-Encoding Functions}

Our event emulation scheme (\cref{alg:event-camera}) relies on the SPAD's response curve to encode scene brightness, which is a non-linear and non-saturating response of the form $$1 - \exp(-\alpha \Phi(\vx, t)),$$ where $\alpha = \eta t_\text{exp}$ and assuming negligible dark count rate (DCR). Current event cameras typically use a logarithmic response to encode scene brightness. This can also be utilized to emulate events from photon-cubes by setting $h$ (as described in \cref{eq:SPAD_event}) to be the log-MLE function: $$h(\mu) = \log\left( -\frac{\log(1- \mu)}{\eta t_\text{exp}}\right).$$ However, a log-response suffers from underflow issues, particularly at low-light scenarios as seen in Supp. Fig. \ref{supl-fig:response-curve}.

\input{supplementary_figures/response_curve.tex}

\subsection*{SoDaCam Flexibility and SPAD-Events}

Here are a few benefits of the SoDaCam approach for event-based imaging:

\begin{itemize}
    \item \textbf{Direct access to intensity information.} By computing a sum image, SoDaCam makes intensity frames that are spatially- and temporally-aligned with the generated event stream available. This precludes the need for multiple devices, which often require careful alignment and calibration.
    \item Further, the intensity frames obtained via the sum-image feature the SPAD's imaging capabilities, i.e., such intensity frames feature a high dynamic range and can be utilized in low-light imaging scenarios. This is in contrast to dynamic active vision sensors (DAVIS) \cite{DAVIS240,Chen_2019_CVPR_Workshops}, where a conventional frame, which has limited dynamic range and low-light capabilities compared to SPAD-derived images, can be obtained in addition to the event stream.
    \item \textbf{Computing multiple event-streams simultaneously.} Contrast threshold $\tau$ is an important parameter that controls the sparsity and noise level of generated event stream: small values of $\tau$ can produce potentially noisy event streams that require extensive processing, while large values of $\tau$ can result in very sparse streams with less useful information. With SoDaCam, it is possible to emulate event streams with different values of $\tau$ simultaneously, thereby amortizing these trade-offs. In fact, this can be thought of as analogous to exposure stacks but in the context of event-imaging. Supplementary Figure \ref{suppl-fig:event-vis} \textit{(top row)} shows an example of an `event-image stack'.
    \item \textbf{Per-pixel contrast thresholds.} We can also vary the contrast threshold $\tau$ as a function of pixel location or incident intensity. For instance, we can use a smaller contrast threshold if we have an estimate of the incident intensity with less variance and a higher contrast threshold when there is more variance. We show an example of this in Supp. Fig. \ref{supl-fig:event-algo} \textit{(bottom right)}, where we vary the contrast threshold between $0.35$ and $0.45$ as a linear function of the sample variance of the moving average, $\mu_t(\vx)$.
\end{itemize}

\input{supplementary_figures/event_vis.tex}

%% file: supplementary_figures/event_algo.tex
\begin{figure*}[htp]
    \centering
    \includegraphics[width=\textwidth]{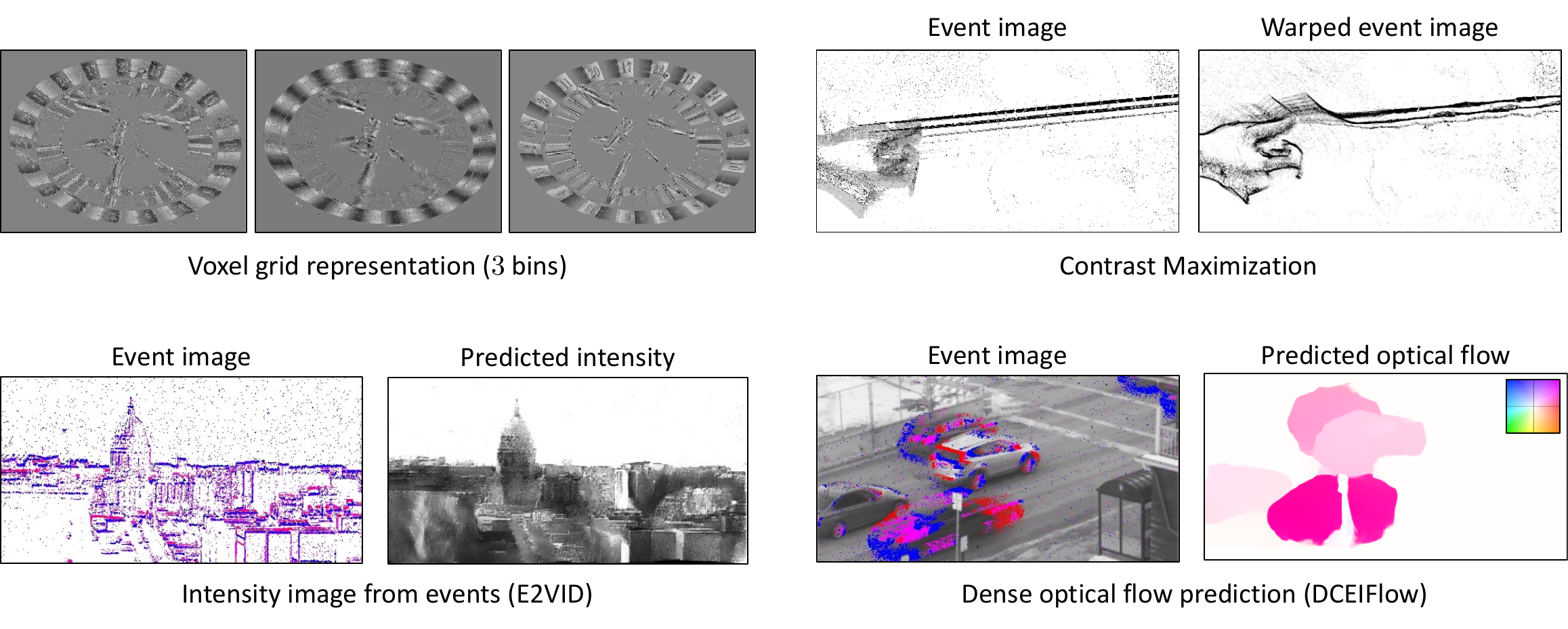} 
    \vspace{-0.15in}
    \tightcaption{\textbf{Compatibility of SPAD-events with existing event-vision algorithms.} The flow field visualization follows \citet{baker2011database}. The photon-cube for the contrast maximization output was obtained from \citet{ma_quanta_2020}.}
    \label{supl-fig:event-algo}
\end{figure*}

%% file: supplementary_figures/response_curve.tex
\begin{figure*}[htp]
    \centering
    \includegraphics[width=0.49\textwidth]{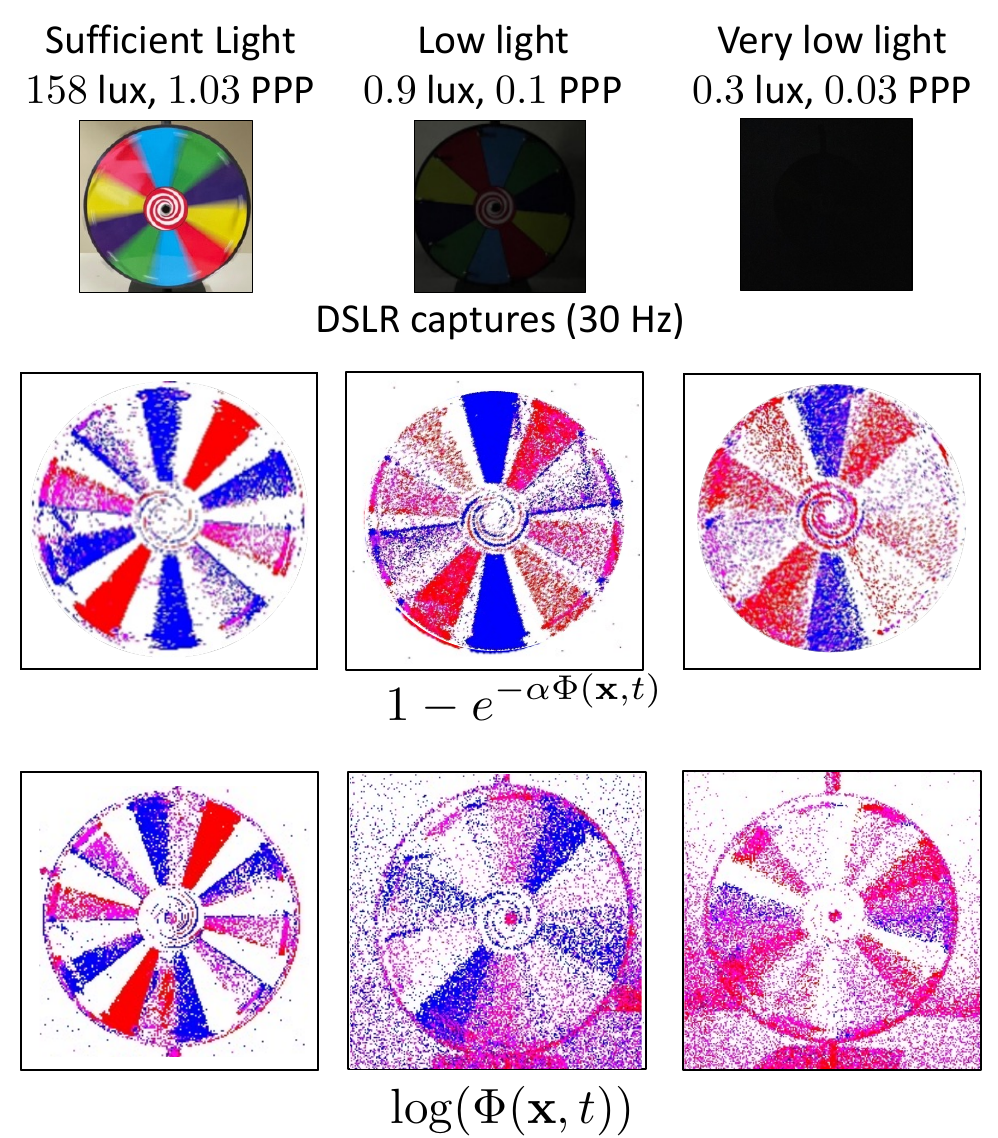} 
    \caption{\textbf{Comparison of brightness encoding functions.} While the log-MLE is comparable to using the SPAD's response curve at ambient light levels, at low flux levels the underflow issues associated with the log function occur. Here, $\alpha$ denotes a sensor-determined and flux-independent constant.}
    \label{supl-fig:response-curve}
\end{figure*}

%% file: supplementary_figures/event_vis.tex
\begin{figure*}[tp]
    \centering
    \includegraphics[width=0.6\textwidth]{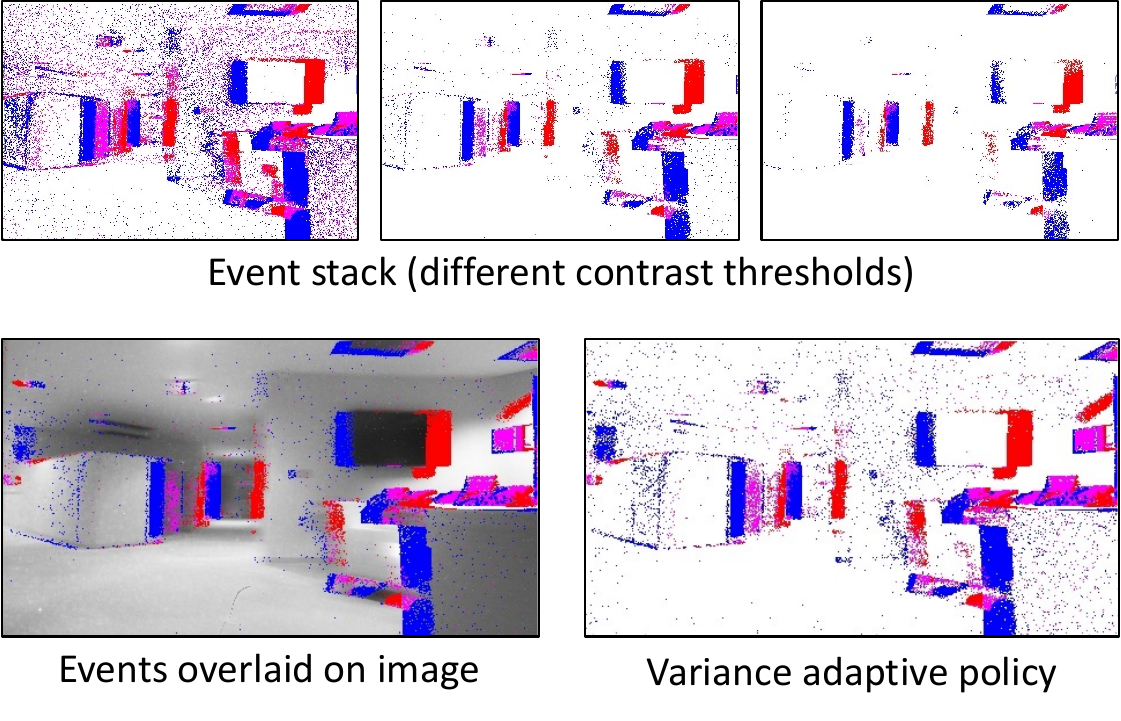}  
    \tightcaption{\textbf{Flexible event-based imaging}. \textit{(top)} An `event-stack' that employs increasing contrast thresholds, $\tau=0.35, 0.4, 0.45$. \textit{(bottom)} We can also output frames with aligned events, and use event generation policies that may not be trivial to realize in hardware---such as varying the contrast threshold $\tau$ as a linear function of the sample variance.}
    \label{suppl-fig:event-vis}
\end{figure*}

%% file: supplementary_sections/motion_projection.tex
\subsection*{Pseudocode for Emulating Motion Cameras}

\Cref{alg:motion-camera} provides the pseudocode for emulating sensor motion from a photon-cube, where the sensor's trajectory is determined by the discretized function $\vr$. At each time instant $t$, we shift bit-planes by $\vr(t)$ and accumulate them in $\cI_\text{shift}$. For pixels that are out-of-bounds, no accumulation is performed. For this reason, the number of summations that occur varies spatially across pixel locations $\vx$---we normalize the emulated shift-image by the number of pixel-wise accumulations $N(\vx)$ to account for this. The function $\vr$ can be obtained by discretizing any smooth $2$D trajectory: by either rounding up or dithering, or by using a discrete line-drawing algorithm \cite{bresenham1965algorithm}.

\begin{algorithm}
\caption{Motion Camera Emulation}\label{alg:motion-camera}
\begin{algorithmic}
\Require {Photon-cube $B_t(\vx)$ \\
Discretized trajectory $\vr(t)$ \\
Pixel locations $\mathcal{X}$\\
Total bit-planes $T$}
\Ensure {$\cI_\text{shift}(\vx)$}
\Function{MotionCameraEmulation}{$B_t(\vx)$, $\vr$}
\State $\cI_\text{shift}(\vx) \gets 0,\, \forall \, \vx$
\For{$\vx \in \cx$}
\State Normalizer, $N(\vx) \gets 0$
\For {$1 \leq t \leq T$}
\If{$\vx + \vr(t) \in \cx$}
\State $N(\vx) \gets N(\vx) + 1$
\State $\cI_\text{shift}(\vx) \gets \cI_\text{shift}(\vx) + B_t(\vx + \vr(t))$
\EndIf
\EndFor
\If{$N(\vx) > 0$}
\State $\cI_\text{shift}(\vx) \gets \cI_\text{shift}(\vx) / N(\vx)$
\EndIf
\EndFor\\
\Return $\cI_\text{shift}(\vx)$
\EndFunction
\end{algorithmic}
\end{algorithm}

As described in \cref{sec:motion-projection}, we consider two trajectories: linear and parabolic. Linear trajectories are parameterized by their slope 
$$ \vr (t) = v \left(t - \frac{T}{2}\right)\uvec{p},$$ 
where $v$ is the object velocity, $\uvec{p}$ is a unit vector that describes the trajectory's direction, and $T$ is the total number of bit-planes. 
Parabolic trajectories are parameterized by their maximum absolute slope, $v_\text{max}$ $$\vr(t) = \frac{v_\text{max}}{T}\left(t - \frac{T}{2}\right)^2 \uvec{p}.$$To prevent tail-clipping, which are image artifacts introduced by the finite extent of the parabolic integration, it is important to choose $v_\text{max}$ to be sufficiently higher than the velocity of objects in the scene. Both linear and parabolic trajectories have a zero at $t=T/2$---which allows blending multiple linear projections without any pixel alignment issues.

\subsection*{Blending Multiple Linear Projections}

As shown in \cref{fig:motion-recons}, randomly sampling multiple linear projections (seen in Supp. Fig. \ref{supl-fig:motion-stack-blending} \textit{(left column)}) can provide motion compensation when only the motion direction, and not the exact extent of motion, is known. To blend these projections, in addition to the randomly sampled linear projections, we also compute two short exposures using bit-planes at the beginning and end of the photon-cube. For the scenes shown in \cref{fig:motion-recons}, we used the first $200$ and the last $200$ bit-planes to emulate short exposures. We then use RAFT \cite{teed2020raft} to predict optical flow between the two short exposures---which can be used to select the linear projection that can best compensate motion as a function of the pixel location (as seen in Supp. Fig. \ref{supl-fig:motion-stack-blending} \textit{(left column)}). We did not have to perform any spatial smoothing after selecting linear projections, since the optical flow predicted by RAFT was reasonably smooth.

Blending can also be achieved by choosing the least blurred linear projection in a per-pixel manner---similar to how focal stacking is typically achieved. This would however require predicting per-pixel blur kernels or constructing a measure of motion blur. Laplacian filters, which are typically used for focal stacking, do not readily work with motion stacks.

\input{supplementary_figures/motion_stack_blending.tex}

%% file: supplementary_figures/motion_stack_blending.tex
\begin{figure*}[htp]
    \centering
    \includegraphics[width=\textwidth]{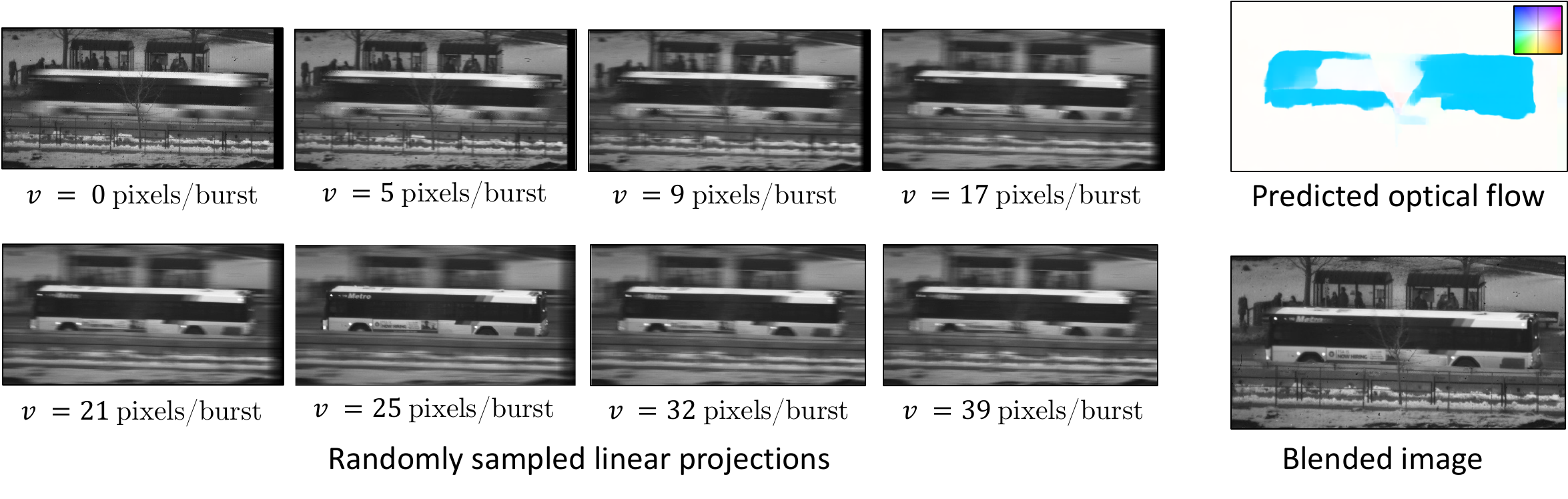} 
    \vspace{-0.15in}
    \tightcaption{\textbf{An example of motion stack blending.} \textit{(left)} We sample $8$ linear projections randomly along the road's orientation such that their ensuing pixel displacements are uniformly between $0$ and $40$ pixels. \textit{(right)} We then blend them using the optical flow field that is predicted between two short exposures computed from the same photon-cube. The flow field visualization follows \citet{baker2011database}.}
    \label{supl-fig:motion-stack-blending}
\end{figure*}

%% file: supplementary_sections/experimental_setup.tex
\subsection*{Cameras and Sensory Arrays Used}

We used the following imagers for our experiments described in \cref{sec:swiss-spad2,sec:comp-real-hardware}:
\begin{itemize}
    \item \textbf{SwissSPAD2 array} a $512 \times 512$ SPAD array that can be operated at a maximum frame rate of $97$ kHz. We operate the SwissSPAD2 in its `half-array' mode: utilizing one of two sub-arrays, with a resolution of $512 \times 256$ pixels. The SPAD pixels have a pixel pitch of $16.4$ $\mu$m and a low fill factor of $10$\%, owing to the lack of microlenses in the prototype.
    \item \textbf{Prophesee EVK4 event camera}, which is a state-of-the-camera commercial event camera featuring a sensor resolution of $1280 \times 720$ pixels, pixel pitch of $4.86$ $\mu$m and a fill-factor of $>77$\%.
    \item \textbf{Photron infinicam} a conventional high-speed camera that can stream acquisition over USB-C at a resolution of $1246 \times 1024$ pixels and $1$ kHz frame-rate. For higher frame rates, it is necessary to reduce the number of rows that are read out---for the example, we use a resolution of $1246 \times 240$ pixels to obtain acquisition at $4$ kHz in \cref{fig:infinicam-comp}.
\end{itemize}

\input{supplementary_figures/camera_setup.tex}
\vspace{-0.1in}

\subsection*{Removing Hot Pixels}

A few SPAD pixels (around $5$\% of the total pixels in our prototype) have extremely high dark current rate and therefore have $B_t(\vx)=1$ almost always. We detect these \textit{hot pixels} by capturing a photon-cube of $100000$ bit-planes in a very dark environment and detecting pixel locations with high photon counts. For video compressive and event imagers, we inpaint projections using OpenCV's implementation of the Telea algorithm \cite{telea2004image}. For motion projections, we do not sum over bit-plane locations that correspond to hot pixels during integration. Further, we remove pixel locations from the hot pixel mask if the motion trajectory provides access to neighboring values that are not hot pixels. We inpaint the motion projection after excluding these points.

\subsection*{Experiment-wise Lens Specifications}

We used C-mount lenses for our experiments with the following focal lengths:
\begin{itemize}
    \item $12$ mm for the comparison to Prophesee EVK4 in \cref{fig:prophesse-comp}. The Prophesee EVK4 and the SwissSPAD2 were used with the same lens specifications.
    \item $16$ mm for the coded exposures shown in \cref{fig:exposure-stack}.
    \item $35$ mm for the spinning casino roulette shown in \cref{fig:event-cam,fig:infinicam-comp}.
    \item $50$ mm for the motion stack shown in \cref{fig:motion-stack} and the traffic scene shown in \cref{fig:motion-recons}.
    \item $75$ mm for the falling die sequence shown in \cref{fig:first-fig}, the measure tape sequence shown in \cref{fig:motion-cam}, and the water splash captured in \cref{fig:comp-video}.
\end{itemize}

%% file: supplementary_figures/camera_setup.tex
\begin{figure*}[htp]
    \centering
    \includegraphics[width=0.6\textwidth]{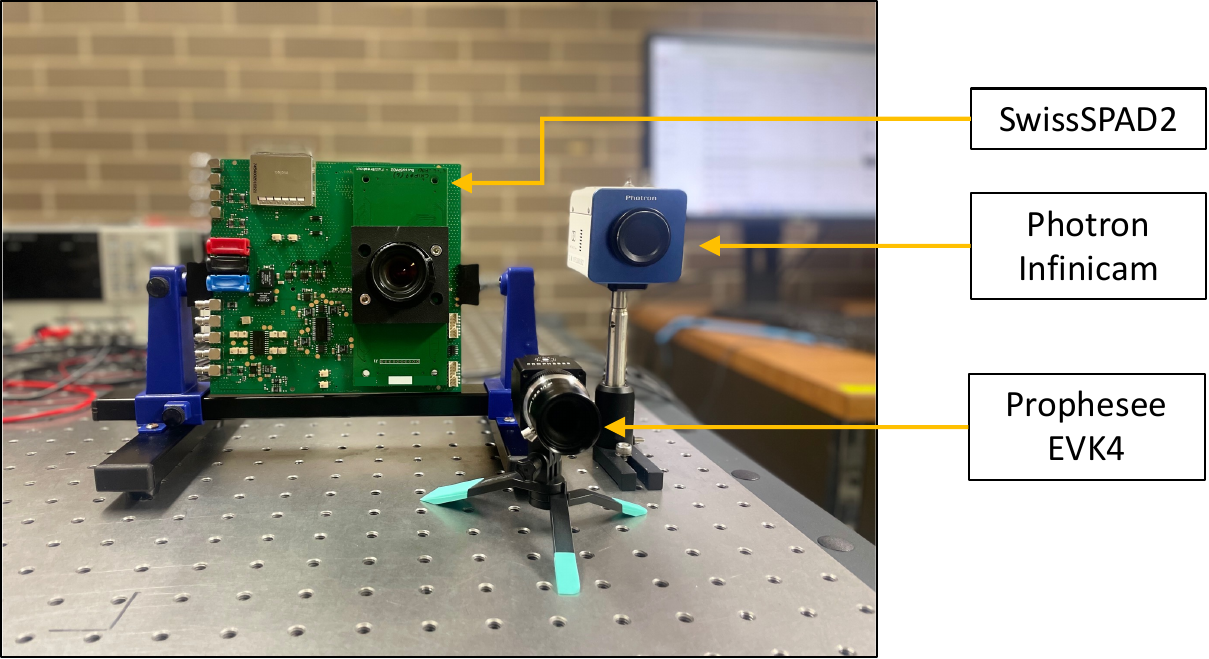} 
    \caption{\textbf{Cameras and sensor arrays used for the experiments described in \cref{sec:swiss-spad2,sec:comp-real-hardware}.} The Infinicam and Prophesee were used for the comparisons made in \cref{fig:infinicam-comp,fig:prophesse-comp} respectively.}
    \label{supl-fig:camera-setup}
\end{figure*}

%% file: supplementary_sections/ultraphase.tex
\subsection*{Processor Description}

The chip consists of a $3 \times 6$ array of processing cores, each of which can interface with $4 \times 4$ SPAD pixels via $3$D stacking. At this point, the $3$D stacking has not been completed, so we interface UltraPhase with the photon-cubes acquired by the SwissSPAD2 \cite{ulku512512SPAD2019} instead. Every core is independent, has $4$ kb of available RAM, and can execute programs of up to 256 instructions in length at a rate of $140$ million instructions per second (MIPS). The system supports a wide range of instructions including, bit-wise operations, 32-bit arithmetic operations, data manipulation and custom inter-core synchronization. For more details, please refer to \citet{ardelean2023computational}.

We implement projections on UltraPhase by using a custom assembly code to program each core separately.
We include the commented assembly code for all three projections in \cref{assembly:vcs,assembly:event,assembly:motion}. To compute multiple projections, we simply run projections sequentially, one bit-plane at a time. Since each projection can be computed significantly faster than the camera frame rate (e.g., $1.678$ ms for video compressive sensing of $40$ Hz readout), this does not bottleneck acquisition. We include the processing time for each projection in \cref{tab:ultraphase}.

\subsection*{Measuring Bandwidth}

We assume that the outputs for sum, video compressive and motion projections have $12$-bit depth. For event cameras, we assume that each event consists of $18$-bits----$9$-bits to encode the pixel location ($\ceil{\log_2(12 \times 24)}$), $8$ bits to represent the timestamp (corresponding to the bit-plane index where the event was triggered), and $1$-bit to encode polarity. We then measure readout on a $12 \times 24$ region-of-interest (RoI) of the falling die sequence that was acquired using the SwissSPAD2. \Cref{tab:ultraphase} lists the readout bandwidth for each projection.

\subsection*{Measuring Power}

The power consumption of UltraPhase is comprised of compute power and readout power. For compute power, the chip was characterized by executing instructions corresponding to each projection in an infinite loop and measuring its average power consumption. As an upper bound, we assumed the maximum possible power consumption for operations that involved reading and writing to the RAM. This measured power consumption was then scaled by the duty cycle of each projection---which is the ratio of the time required to process a bit-plane to the exposure time of each bit-plane.

For readout power, we consider a conventional digital interface at $3.3$ V with a load of $7$ pF operating at the specified bandwidth, amounting to $54$ nanowatts for each kilobit readout (nW/kbps)---this is similar, for instance, to the USB interface utilized by the SwissSPAD2.

\Cref{tab:ultraphase} provides the processing power, readout power and the total power for each projection. Clearly, processing requires an order of magnitude (or more) lesser power than readout, which explains how computing photon-cube projections results in reduced sensor power consumption.

\input{tables/ultraphase}

\subsection*{Comparison to CMOS Sensors} 

In addition to compute and readout power quantified in the previous section, a computational SPAD also consumes power to detect photons. By incorporating this photon-detection power, we can provide a rough comparison of SoDaCam projections to CMOS sensors. The photon-detection dissipation depends on the number of photon detections, and hence varies with the light level---for the SwissSPAD2, this is measured to be $< 1$mW in the dark and \tildeNice$62$ mW in indoor lighting~\cite{ulku512512SPAD2019}.
To estimate compute and readout power, we linearly scale the measurements presented in \cref{tab:ultraphase} for an array of $512 \times 256$ pixels.
We note that this is a conservative estimate since UltraPhase is not designed to be a low-power device.

As seen in \cref{tab:ultraphase}, under ambient lighting, the power consumption of our emulated cameras is higher than conventional CMOS cameras; while in low light, the SPAD consumes lesser power owing to fewer photon detections.
We remark that without the bandwidth reduction facilitated by photon-cube projections, SPADs are at a considerable disadvantage compared to their CMOS counterparts. 
Finally, we provide a comparison against high-speed CMOS cameras, which can also be used to obtain photon-cube projections, albeit with a read noise penalty (\cref{sec:comp-real-hardware}), and higher power consumption.

\input{tables/cmos_comparison_suppl}

\subsection*{Visualization of Projections}

Since UltraPhase is a low-resolution sensor-processor ($12 \times 24$ pixels), we visualize projections by repeating computations in a tiled manner to cover a region-of-interest (RoI) of $60 \times 60$ pixels. Supplementary Figure \ref{supl-fig:ultraphase-vis} shows the visualization of an event camera emulated on UltraPhase. To provide more context, we include the CPU visualization of the entire SwissSPAD2 event-frame. We also verified that the outputs of UltraPhase were identical to CPU-run outputs by computing the RMSE between event frames that result from UltraPhase computations and CPU computations (see error map in Supp. Fig. \ref{supl-fig:ultraphase-vis}).

\input{supplementary_figures/ultraphase_vis.tex}

\clearpage
\input{supplementary_sections/assembly_video_compressive.tex}

\clearpage
\input{supplementary_sections/assembly_event.tex}

\clearpage
\input{supplementary_sections/assembly_motion.tex}

%% file: tables/ultraphase.tex
\begin{table}[h]
  \centering
  \caption{\textbf{Power and bandwidth benchmarks} when computing photon-cube projections on UltraPhase, a $24 \times 12$ array, at $40$ Hz readout. We compare computing projections to reading out the entire photon-cube. We report the processing time, the readout bandwidth, and the compute and readout power for each projection.\label{tab:ultraphase}}
\begin{tabular}{@{}lccccc@{}}
\toprule

&
\multicolumn{1}{c}{Processing time $\downarrow$} &
\multicolumn{1}{c}{Bandwidth $\downarrow$}    &
\multicolumn{3}{c}{Power $\downarrow$}    \\ 
\cmidrule(lr){4-6}

&
\multicolumn{1}{c}{(ms)} &
\multicolumn{1}{c}{(kbps)}     &
\multicolumn{1}{c}{Processing ($\mu$W)} &
\multicolumn{1}{c}{Readout ($\mu$W)} &
\multicolumn{1}{c}{Total ($\mu$W)} \\
\midrule

12-bit sum image & $0.981$ & $135$ & $0.3$ & $7.29$ & $7.6$\\
Snapshot compressive & $1.678$ & $135$ & $3.0$ & $7.29$ & $10.3$\\
Motion projection & $1.096$ & $135$ & $1.3$ & $7.29$ & $8.6$\\
Event camera & $9.817$ & $101.25$ & $2.4$ & $5.83$ & $8.2$\\
\midrule
Three projections & $12.591$ & $405$ & $6.7$ & $21.87$ & $28.6$\\
\midrule
Photon-cube readout & $0.007$ & $28125$ & $5.4 \times 10^{-3}$ & $1518.8$ & $1518.8$ \\
\bottomrule
\end{tabular}
\end{table}

%% file: tables/cmos_comparison_suppl.tex
\begin{table}[t]
    \caption{\textbf{Power consumption} of SoDaCam versus conventional cameras (in mW), estimated for $512 \times 256$ pixels at $40$ Hz readout. 
    CMOS estimates assume the usage of column-parallel ADCs~\cite{SnoeijJSSC2007}.}
  \centering
    \begin{tabular}{@{}lcccccc@{}}
    \toprule    
    &
    \multicolumn{2}{c}{Photon detection}    &
    \multirow{2}{*}{Compute}    &
    \multirow{2}{*}{Readout}    &
    \multicolumn{2}{c}{Total}    \\ 

    \cmidrule(lr){2-3}
    \cmidrule(lr){6-7}
    &
    Dark & Ambient &
    & &
    Dark & Ambient \\
    \midrule

    \makecell[l]{Photon-cube \\readout} & $1$ & $62$ & - & $690$ & $691$ & $752$ \\
    \cmidrule{2-7}
    Sum-image & $1$ & $62$ & $0.3$ & $4.5$ & $5.8$ & $66.8$ \\
    VCS & $1$ & $62$ & $1.3$ & $4.5$ & $6.8$ & $67.8$ \\
    Motion proj. & $1$ & $62$ & $0.7$ & $4.5$ & $6.2$ & $67.2$\\
    Event camera & $1$ & $62$ & $1$ & $3.6$ & $5.6$ & $66.6$\\
    Three proj.(s) & $1$ & $62$ & $3$ & $13.5$ & $17.5$ & $78.5$\\
    \midrule

    CMOS @ 40 FPS & \multicolumn{2}{c}{$\tildeNice 10$--$25$} & - & $4.5$ & \multicolumn{2}{c}{$\tildeNice 15$--$30$} \\
    \cmidrule{2-7}
    CMOS @ 4k FPS & \multicolumn{2}{c}{$\tildeNice 600$--$2500$} & - & $450$ & \multicolumn{2}{c}{$\tildeNice 1000$--$3000$} \\
    \bottomrule
    \end{tabular}
\end{table}

%% file: supplementary_figures/ultraphase_vis.tex
\begin{figure*}[htp]
    \centering
    \includegraphics[width=0.85\textwidth]{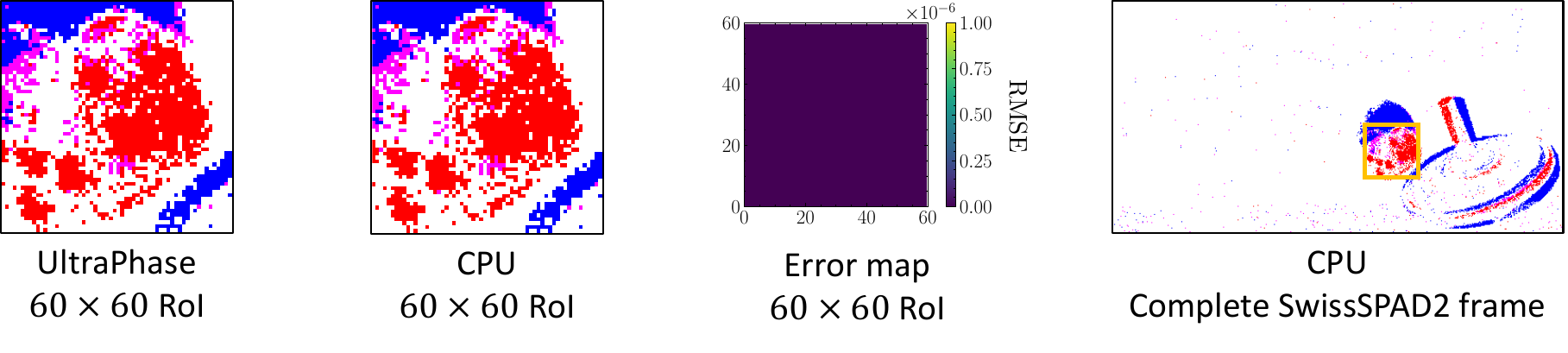} 
    \tightcaption{\textbf{Event camera computed using UltraPhase,} on $2500$ bit-planes of the falling die sequence. For visualization purposes, we run computations on UltraPhase in a tiled manner so as to cover a RoI of $60\times 60$ pixels. We compare this to CPU-run outputs of the same RoI and verify that they are identical. For context, we highlight this RoI using a bounding box on the CPU-run event-frame that has a resolution of $256 \times 512$ pixels. The event simulation parameters used were $\tau=0.45, \beta=0.95, T_0 = 80$. }
    \label{supl-fig:ultraphase-vis}
\end{figure*}

%% file: supplementary_sections/assembly_video_compressive.tex
\begin{code}
\caption{\textbf{Custom assembly code for implementing video compressive sensing on UltraPhase.} Here, we consider computing one compressive snapshot that is multiplexed by $16$ binary random masks.\label{assembly:vcs}}
\vspace{-0.1in}
\begin{minted}[breaklines,frame=lines,fontsize=\footnotesize]{nasm}
--RAM(64..127) stores the 64 subframe compression code masks as one bit per pixel in byte1 and byte0
--The output is available in RAM(0..15)
--CtrlOut is strobed after every binary frame

#define pixelValue 0				--register R0 is used for the pixel values
#define frameIdx 1					--register R1 is used as a counter for the current binary frame
#define subFrameIdx 2				--register R2 is used as a counter for the current subFrame
#define mask 3						--register R3 is used to store the compression codes for the current subframe
#define aux0 4						--register R4 is used for 
#define aux1 5						--register R5 is used for 
#define CtrlOut 0b10000				--address for external trigger signal
#define toMask 0b001000				--address used to fetch to R3 (mask)

0: LOAD subFrameIdx, $64, 0			--initialize subFrame pointer to 0 + 64 (RAM offset)
1: LOAD subFrameIdx, $0, 1
2: LOAD frameIdx, $16, 0			--initialize frameIdx pointer to 16 i.e. number of bit_planes_per_subframe
3: LOAD frameIdx, $0, 1

4: GETP 5, 1, 0 					--get pixel data and store it in R0 (pixelValue)
5: FETCH @subFrameIdx, toMask, 0	--get the appropriate mask
6: AND pixelValue, mask, pixelValue	--apply mask i.e. multiply with code[subframe]
7: CALL 50							--accumulate pixels
8: TELL CtrlOut, 0					--strobe CtrlOut	

-----------------------------------------------------
-- update indexes
9: OR aux0, aux0, aux0				--clear flags
10: SUBC frameIdx, $1 				--decrement frameIdx
11: JUMPZ 13						--check if we finished with all the binary planes per subframe
12: JUMP 4							--if not, then move on
13:  OR aux0, aux0, aux0			--clear flags
14: ADDC subFrameIdx, $1			--move to next subFrame
15: JUMP 2 							--reset frameIdx and continue

-----------------------------------------------------
-- this subroutine will accumulate the pixel values from pixelValue
50: LOAD aux0, $16, 0				--set R4 (aux0) to 16
51: LOAD aux0, $0, 1
52: LOAD aux1, $0, 0				--set R5 (aux1) to 0
53: LOAD aux1, $0, 1
54: OR aux0, aux0, aux0				--clear flags
55: SR0 pixelValue, pixelValue 		--extract one pixel
56: ADDC (aux0), aux1, (aux0)		--increment if pixel is 1
57: SUBC aux0, $1					--decrement counter
58: JUMPNZ 54						--repeat until done with all 16 pixels
59: RET
\end{minted}
\end{code}

%% file: supplementary_sections/assembly_event.tex
\begin{code}
\caption{\textbf{Custom assembly code for implementing event cameras on UltraPhase.} \label{assembly:event}}
\vspace{-0.1in}
\begin{minted}[breaklines,frame=lines,fontsize=\footnotesize]{nasm}
--The core will strobe CtrlOut every time an evet took place and the SoC needs to read RAM(0) to get it

#define toaux0 0b000010
#define toaux1 0b000100
#define pixelValue 0	--register R0 is used for the binary pixel values
#define aux0 1			--register R1 is used for misc
#define aux1 2			--register R2 is used for misc
#define Bpointer 3		--register R3 is used as a pointer to the reference_average stored in RAM(17..32) 
#define Apointer 4		--register R4 is used as a pointer to the current_average stored in RAM(1..16) 
#define pixelIdx 5		--register R5 is used for the current pixel index
#define CtrlOut 0b10000				--address for external trigger signal
#define NOut	0b01000				--address for north trigger signal
#define decayAddress 127			--RAM address 127 stores the exponential_decay
#define decayComplementAddress 126	--RAM address 126 stores the value for 1-exponential_decay
#define intervalAddress 125			--RAM address 125 stores the initial_interval
#define thresholdAddress 124		--RAM address 124 storesthe contrast_threshold
#define frameIdxAddress 123			--RAM address 123 stores the frame_index counter

0: GETP 5, 1, 0 								--get pixel data and store it in R0 (pixelValue)
1: LOAD pixelIdx, $0, 1							--set R5 (pixelIdx) to 16 to use as counter for the pixels
2: LOAD pixelIdx, $16, 0
3: LOAD Apointer, $0, 1							--initialize pointers
4: LOAD Apointer, $1, 0
5: LOAD Bpointer, $0, 1
6: LOAD Bpointer, $17, 0
 
7: FETCH @decayAddress, toaux0, 0				--get decay constant from RAM and store it into R1 (aux0)
8: FETCH @decayComplementAddress, toaux1, 0		--get (1 - decay) constant from RAM and store it into R2 (aux1)

9: MUL (Apointer), aux0, (Apointer)				-- current_average[pixelIdx] = exponential_decay * current_average[pixelIdx]
10: SRX (Apointer), (Apointer)					-- 8 fractional bit multiplication
11: SRX (Apointer), (Apointer)
12: SRX (Apointer), (Apointer)
13: SRX (Apointer), (Apointer)
14: SRX (Apointer), (Apointer)
15: SRX (Apointer), (Apointer)
16: SRX (Apointer), (Apointer)
17: SRX (Apointer), (Apointer)

18: SR0 pixelValue, pixelValue					--shift R0 (pixelValue) to the right and pad with 0; the pixel bit is loaded into the carry flag 
19: JUMPNC 21 									--current_average[pixelIdx] = current_average[pixelIdx] + (1-exponential_decay) * 0
20: ADD (Apointer), aux1, (Apointer)  			--current_average[pixelIdx] = current_average[pixelIdx] + (1-exponential_decay) * 1
21: FETCH @frameIdxAddress, toaux0, 0	    	--get frame_index from RAM and store it into R1 (aux0)
22: FETCH @intervalAddress, toaux1, 0	    	--get initial_interval from RAM and store it into R2 (aux1)
23: CMP aux0, aux1								--check if frame_index < initial_interval
24: JUMPNC 40									--if not, jump and process 
25: FETCH (Apointer), toaux0, 0					--if yes, get current_average[pixelIdx]
26: STORE aux0, (Bpointer)						--reference_average[pixelIdx] = current_average[pixelIdx]
27: JUMP 70										--GO TO NEXT PIXEL

-- frame_index is larger than initial_interval
40: FETCH (Apointer), toaux0, 0	    			--if not, get current_average[pixelIdx] and store it into R1 (aux0)
41: FETCH (Bpointer), toaux1, 0					--get reference_average[pixelIdx] and store it into R2 (aux1)
42: SUB aux0, aux1, aux0						--diff[pixelIdx] = current_average[pixelIdx] - reference_average[pixelIdx] and store it into R1 (aux0)
43: JUMPNC	60									--diff is positive

-- compare abs(diff) with threshold if diff is negative
44: FETCH @thresholdAddress, toaux1, 0			--get contrast_threshold and store it into R2 (aux1)
45: NEG aux1, aux1								--diff is negative, so make contrast_threshold negative and store it into R2 (aux1)
46: CMP aux0, aux1								--if diff < -contrast_threshold
47: JUMPNC 70									--if not, GO TO NEXT PIXEL
48: NEG pixelIdx, @0   							--RAM(0) = -pixel_index i.e. a negative event
49: ADD (Bpointer), aux1, (Bpointer)			--reference_average += contrast_threshold * (-1)
50: TELL CtrlOut, 0								--strobe CtrlOut to signal an event
51: JUMP 70										--GO TO NEXT PIXEL

-- compare abs(diff) with threshold if diff is positive
60:  FETCH @thresholdAddress, toaux1, 0			--get contrast_threshold and store it into R2 (aux1)
61: CMP aux1, aux0								--if diff > contrast_threshold
62: JUMPNC 70									--if not, GO TO NEXT PIXEL
63: STORE pixelIdx, @0							--RAM(0) = pixel_index i.e. a positive event
64: ADD (Bpointer), aux1, (Bpointer)			--reference_average += contrast_threshold * 1
65: TELL CtrlOut, 0								--strobe CtrlOut to signal an event
66: JUMP 70										--GO TO NEXT PIXEL

--GO TO NEXT PIXEL
70: OR pixelValue, pixelValue, pixelValue 		--clear flags
71: ADDC Apointer, $1							--increment Apointer
72: ADDC Bpointer, $1							--increment Bpointer
73: SUBC pixelIdx, $1							--decrement pixel counter
74: JUMPNZ 7									--if not done with pixels, go to next one
75: FETCH @frameIdxAddress, toaux0, 0			--if done with pixels, increment frame counter and get new pixel values
76: TELL NOut, 0								--strobe NOut to signal a new exposure
77: ADDC aux0, $1
78: STORE aux0, @frameIdxAddress
79: JUMP 0
\end{minted}
\end{code}

%% file: supplementary_sections/assembly_motion.tex
\begin{code}
\caption{\textbf{Custom assembly code for implementing motion projections on UltraPhase.} Without loss of generality, we consider a linear projection along the horizontal direction.\label{assembly:motion}}
\vspace{-0.1in}
\begin{minted}[breaklines,frame=lines,fontsize=\footnotesize]{nasm}
--The projection is available in RAM(0..15)
-- CtrlOut is strobed after every frame

#define Xshift 0		--register R0 is used for the horizontal shift
#define timestep 1		--register R1 is used for the current timestep
#define origPixels 2	--register R2 is used for the current core's pixels
#define shiftPixels 3	--register R3 is used for the shifted pixels 
#define aux0 4			--register R4 is used for misc
#define aux1 5			--register R5 is used for misc
#define CtrlOut 0b10000		    --address for external trigger signal
#define shiftL3neighAddr 127	--RAM(127) stores the 0b0111_0111_0111_0111 mask
#define shiftL3currAddr 126		--RAM(126) stores the 0b1000_1000_1000_1000 mask
#define shiftL2neighAddr 125	--RAM(125) stores the 0b0011_0011_0011_0011 mask
#define shiftL2currAddr 124		--RAM(124) stores the 0b1100_1100_1100_1100 mask
#define shiftL1neighAddr 123	--RAM(123) stores the 0b0001_0001_0001_0001 mask
#define shiftL1currAddr 122		--RAM(122) stores the 0b1110_1110_1110_1110 mask

0: LOAD Xshift, $0xFFF8, 0	--load -8 into Xshift as initial value
1: LOAD Xshift, $0xFFFF, 1
2: CALL 50 					--get the correct pixel values according to Xshift
3: CALL 30 					--accumulate pixels
4: CALL 20 					--advance time
5: TELL CtrlOut, 0			-- strobe CtrlOut to signal a new frame
6: JUMP 2    				--repeat

-----------------------------------------------------
-- this subroutine will advance timestep and update Xshift
20: OR timestep, timestep, timestep	--clear flags
21: ADDC timestep, $1				--increment timestep
22: OR timestep, timestep, aux0		--copy timestep to aux0
23: SPLIT aux0, 5, 0b010000			--timestep = timestep/1024
24: SR0 aux0, aux0					
25: SR0 aux0, aux0					
26: LOAD aux1, $1, 0				--load aux1 with the Xspeed value of 1
27: MAC aux0, aux1, Xshift, Xshift	--Xshift += Xspeed*timestep/2
28: RET

-----------------------------------------------------
-- this subroutine will accumulate the pixel values from shiftPixels
30: LOAD aux0, $16, 0				--set R4 (aux0) to 16
31: LOAD aux0, $0, 1
32: LOAD aux1, $0, 0				--set R5 (aux1) to 0
33: LOAD aux1, $0, 1
34: OR aux0, aux0, aux0				--clear flags
35: SR0 shiftPixels, shiftPixels 	--extract one pixel
36: ADDC (aux0), aux1, (aux0)		--increment if pixel is 1
37: SUBC aux0, $1					--decrement counter
38: JUMPNZ 34						--repeat until done with all 16 pixels
39: RET 

-----------------------------------------------------
-- this subroutine will get pixel values from the core and the correct neighbour based on the Xshift
50: GETP 5, 4, 0 					--get pixel values and store them in R2 (origPixels)
51: OR Xshift, Xshift, aux1			--copy current shift into R5 (aux1)
52: LOAD aux0, $0, 0				--set R4 (aux0) to 0 to use for Xshift comparison
53: LOAD aux0, $0, 1
54: CMP aux1, aux0 					--compare current shift with 0
55: JUMPZ 115						--current shift = 0
56: JUMPC 120						--current shift < 0
57: JUMP 58							--current shift > 0
-------------------------------
-- positive shifts
58: LOAD aux0, $4, 0				--set R4 (aux0) to 4 for current shift comparison
59: LOAD aux0, $0, 1
60: CMP aux0, aux1
61: JUMPC 105			--current shift > 4, read from the neighbour's neighbour
62: JUMPZ 100						--current shift is 4 
63: LOAD aux0, $2, 0				--set R4 (aux0) to 2, for current shift comparison
64: CMP aux0, aux1
65: JUMPC 90 						--current shift is 3
66: JUMPZ 80						--current shift is 2

-- shift is 1
67: SL0 shiftPixels, shiftPixels			--pixels from the neighbour need to be shifted to the left 3 times
68: SL0 shiftPixels, shiftPixels
69: SL0 shiftPixels, shiftPixels
70: OR origPixels, origPixels, aux1 			--save curent pixels in the aux1 variable
71: SR0 aux1, aux1							--pixels from this core need to be shifted to the right once
72: AND shiftPixels, @shiftL3currAddr, shiftPixels	--apply mask to select relevant bits from neighbour
73: AND aux1, @shiftL3neighAddr, aux1			--apply mask to select relevant bits from current core
74: OR shiftPixels, aux1, shiftPixels			--combine to create final pixel values
75: RET

-- shift is 2
80: SL0 shiftPixels, shiftPixels					--pixels from neighbour need to be shifted to the left 2 times
81: SL0 shiftPixels, shiftPixels
82: OR origPixels, origPixels, aux1 				--save curent pixels in the aux1 variable
83: SR0 aux1, aux1									--pixels from this core need to be shifted to the right 2 times
84: SR0 aux1, aux1
85: AND shiftPixels, @shiftL2currAddr, shiftPixels	--apply mask to select relevant bits from neighbour
86: AND aux1, @shiftL2neighAddr, aux1				--apply mask to select relevant bits from current core
87: OR shiftPixels, aux1, shiftPixels				--combine to create final pixel values
88: RET

-- shift is 3
90: SL0 shiftPixels, shiftPixels					--pixels from neighbour need to be shifted to the left once
91: OR origPixels, origPixels, aux1 				--save curent pixels in the aux1 variable
92: SR0 aux1, aux1									--pixels from this core need to be shifted to the right 3 times
93: SR0 aux1, aux1
94: SR0 aux1, aux1
95: AND shiftPixels, @shiftL1currAddr, shiftPixels	--apply mask to select relevant bits from neighbour
96: AND aux1, @shiftL1neighAddr, aux1				--apply mask to select relevant bits from current core
97: OR shiftPixels, aux1, shiftPixels				--combine to create final pixel values
98: RET

-- shift is 4
100: PUTN origPixels, 4			--shift your pixels to the left (share pixels with neighbour)
101: SAVEN 1					--save pixels from rigth neighbour
102: GETN 0, 8, 0				--get pixels from rigth neighbour and store in shiftPixels
103: RET 

-- shift is > 4
105: PUTN origPixels, 4			--shift your pixels to the left (share pixels with neighbour)
106: SAVEN 1					--save pixels from right neighbour
107: GETN 0, 8, 0				--get pixels from right neighbour and store in shiftPixels
108: OR aux1, aux1, aux1		--clear Carry flag
109: SUBC aux1, $4				--subtract 4 from current shift because we read from a neighbour
110: JUMP 58

-------------------------------
-- shift is zero
115: OR origPixels, origPixels, shiftPixels  --the shift is zero, so keep the pixels
116: RET 

-------------------------------
-- negative shifts
120: LOAD aux0, $FFFC, 0		--set R4 (aux0) to -4 to use for current shift comparison
121: LOAD aux0, $0xFFFF, 1
122: CMP aux1, aux0
123: JUMPC 170					--current shift < -4 so we need to read from neighbour's neighbour
124: JUMPZ 160					--current shift is -4 
125: LOAD aux0, $FFFE, 0		--set R4 (aux0) to -2 to use for current shift comparison
126: CMP aux1, aux0
127: JUMPC 150 					--current shift is -3
128: JUMPZ 140					--current shift is -2

-- shift is -1
129: SR0 shiftPixels, shiftPixels						--pixels from neighbour need to be shifted to the right 3 times
130: SR0 shiftPixels, shiftPixels
131: SR0 shiftPixels, shiftPixels
132: OR origPixels, origPixels, aux1 					--save curent pixels in the aux1 variable
133: SL0 aux1, aux1										--pixels from this core need to be shifted to the left once
134: AND shiftPixels, @shiftL1neighAddr, shiftPixels	--apply mask to select relevant bits from neighbour
135: AND aux1, @shiftL1currAddr, aux1					--apply mask to select relevant bits from current core
136: OR shiftPixels, aux1, shiftPixels					--combine to create final pixel values
137: RET

-- shift is -2
140: SR0 shiftPixels, shiftPixels						--pixels from neighbour need to be shifted to the right 2 times
141: SR0 shiftPixels, shiftPixels
142: OR origPixels, origPixels, aux1 					--save curent pixels in the aux1 variable
143: SL0 aux1, aux1										--pixels from this core need to be shifted to the left 2 times
144: SL0 aux1, aux1
145: AND shiftPixels, @shiftL2neighAddr, shiftPixels	--apply mask to select relevant bits from neighbour
146: AND aux1, @shiftL2currAddr, aux1					--apply mask to select relevant bits from current core
147: OR shiftPixels, aux1, shiftPixels					--combine to create final pixel values
148: RET

-- shift is -3
150: SR0 shiftPixels, shiftPixels						--pixels from neighbour need to be shifted to the right once
151: OR origPixels, origPixels, aux1 					--save curent pixels in the aux1 variable
152: SL0 aux1, aux1										--pixels from this core need to be shifted to the left 3 times
153: SL0 aux1, aux1
154: SL0 aux1, aux1
155: AND shiftPixels, @shiftL3neighAddr, shiftPixels	--apply mask to select relevant bits from neighbour
156: AND aux1, @shiftL3currAddr, aux1					--apply mask to select relevant bits from current core
157: OR shiftPixels, aux1, shiftPixels					--combine to create final pixel values
158: RET

-- shift is -4
160: PUTN origPixels, 1			--shift your pixels to the right (share pixels with neighbour)
161: SAVEN 4					--save pixels from left neighbour
162: GETN 2, 8, 0				--get pixels from left neighbour and store in shiftPixels
163: RET 

-- shift is < -4
170: PUTN origPixels, 1			--shift your pixels to the right (share pixels with neighbour)
171: SAVEN 4					--save pixels from left neighbour
172: GETN 2, 8, 0				--get pixels from left neighbour and store in shiftPixels
173: OR aux1, aux1, aux1		--clear Carry flag
174: ADDC aux1, $4				--add 4 to current shift because we read from a neighbour
175: JUMP 120
\end{minted}
\end{code}

%% file: main.bbl
\begin{thebibliography}{78}
\providecommand{\natexlab}[1]{#1}
\providecommand{\url}[1]{\texttt{#1}}
\expandafter\ifx\csname urlstyle\endcsname\relax
  \providecommand{\doi}[1]{doi: #1}\else
  \providecommand{\doi}{doi: \begingroup \urlstyle{rm}\Url}\fi

\bibitem[pha()]{phantom_v2640}
Phantom-v2640.
\newblock
  \url{https://www.phantomhighspeed.com/products/cameras/ultrahigh4mpx/v2640}.
\newblock Accessed: 2023-01-28.

\bibitem[Adelson et~al.(1991)Adelson, Bergen, et~al.]{adelson1991plenoptic}
E.~H. Adelson, J.~R. Bergen, et~al.
\newblock The plenoptic function and the elements of early vision.
\newblock \emph{Computational models of visual processing}, 1\penalty0
  (2):\penalty0 3--20, 1991.

\bibitem[Agrawal et~al.(2010)Agrawal, Veeraraghavan, and
  Raskar]{agrawal2010reinterpretable}
A.~Agrawal, A.~Veeraraghavan, and R.~Raskar.
\newblock Reinterpretable imager: Towards variable post-capture space, angle
  and time resolution in photography.
\newblock In \emph{Computer Graphics Forum}, volume~29, pages 763--772. Wiley
  Online Library, 2010.

\bibitem[Antolovic et~al.(2016)Antolovic, Burri, Bruschini, Hoebe, and
  Charbon]{Antolovic2016}
I.~M. Antolovic, S.~Burri, C.~Bruschini, R.~Hoebe, and E.~Charbon.
\newblock Nonuniformity analysis of a 65-kpixel {CMOS SPAD} imager.
\newblock \emph{IEEE Transactions on Electron Devices}, 63\penalty0
  (1):\penalty0 57--64, 2016.
\newblock \doi{10.1109/TED.2015.2458295}.

\bibitem[Ardelean(2023)]{ardelean2023computational}
A.~Ardelean.
\newblock \emph{Computational Imaging SPAD Cameras}.
\newblock PhD thesis, École polytechnique fédérale de Lausanne, 2023.

\bibitem[Boukhayma et~al.(2016)Boukhayma, Peizerat, and Enz]{readout_noise}
A.~Boukhayma, A.~Peizerat, and C.~Enz.
\newblock A sub-0.5 electron read noise {VGA} image sensor in a standard {CMOS}
  process.
\newblock \emph{IEEE Journal of Solid-State Circuits}, 2016.

\bibitem[Canny(1986)]{Canny1986}
J.~Canny.
\newblock A computational approach to edge detection.
\newblock \emph{IEEE Transactions on Pattern Analysis and Machine
  Intelligence}, PAMI-8\penalty0 (6):\penalty0 679--698, 1986.
\newblock \doi{10.1109/TPAMI.1986.4767851}.

\bibitem[Carey et~al.(2013)Carey, Lopich, Barr, Wang, and Dudek]{scamp_2013}
S.~J. Carey, A.~Lopich, D.~R. Barr, B.~Wang, and P.~Dudek.
\newblock A 100,000 fps vision sensor with embedded {535GOPS/W 256×256 SIMD}
  processor array.
\newblock In \emph{2013 Symposium on VLSI Circuits}, pages C182--C183, 2013.

\bibitem[Chen et~al.(2017)Chen, Carey, and Dudek]{chen2017feature}
J.~Chen, S.~J. Carey, and P.~Dudek.
\newblock Feature extraction using a portable vision system.
\newblock In \emph{IEEE/RSJ Int. Conf. Intell. Robots Syst., Workshop
  Vis.-based Agile Auton. Navigation UAVs}, volume~2, 2017.

\bibitem[Chen and Guo(2019)]{Chen_2019_CVPR_Workshops}
S.~Chen and M.~Guo.
\newblock Live demonstration: Celex-v: A 1m pixel multi-mode event-based
  sensor.
\newblock In \emph{Proceedings of the IEEE/CVF Conference on Computer Vision
  and Pattern Recognition (CVPR) Workshops}, June 2019.

\bibitem[Dalal and Triggs(2005)]{dalal2005histograms}
N.~Dalal and B.~Triggs.
\newblock Histograms of oriented gradients for human detection.
\newblock In \emph{2005 IEEE computer society conference on computer vision and
  pattern recognition (CVPR'05)}, volume~1, pages 886--893. Ieee, 2005.

\bibitem[Debevec and Malik(1997)]{debevec2008recovering}
P.~E. Debevec and J.~Malik.
\newblock Recovering high dynamic range radiance maps from photographs.
\newblock In \emph{Proceedings of the 24th Annual Conference on Computer
  Graphics and Interactive Techniques}, page 369–378. ACM
  Press/Addison-Wesley Publishing Co., 1997.
\newblock \doi{10.1145/258734.258884}.
\newblock URL \url{https://doi.org/10.1145/258734.258884}.

\bibitem[Della~Rocca et~al.(2020)Della~Rocca, Mai, Hutchings, Al~Abbas,
  Buckbee, Tsiamis, Lomax, Gyongy, Dutton, and Henderson]{della2020128}
F.~M. Della~Rocca, H.~Mai, S.~W. Hutchings, T.~Al~Abbas, K.~Buckbee,
  A.~Tsiamis, P.~Lomax, I.~Gyongy, N.~A. Dutton, and R.~K. Henderson.
\newblock A 128$\times$ 128 {SPAD} motion-triggered time-of-flight image sensor
  with in-pixel histogram and column-parallel vision processor.
\newblock \emph{IEEE Journal of Solid-State Circuits}, 55\penalty0
  (7):\penalty0 1762--1775, 2020.

\bibitem[Finateu et~al.(2020)Finateu, Niwa, Matolin, Tsuchimoto, Mascheroni,
  Reynaud, Mostafalu, Brady, Chotard, LeGoff, Takahashi, Wakabayashi, Oike, and
  Posch]{Finateau2020Prophesee}
T.~Finateu, A.~Niwa, D.~Matolin, K.~Tsuchimoto, A.~Mascheroni, E.~Reynaud,
  P.~Mostafalu, F.~Brady, L.~Chotard, F.~LeGoff, H.~Takahashi, H.~Wakabayashi,
  Y.~Oike, and C.~Posch.
\newblock 5.10 a 1280×720 back-illuminated stacked temporal contrast
  event-based vision sensor with 4.86µm pixels, {1.066GEPS} readout,
  programmable event-rate controller and compressive data-formatting pipeline.
\newblock In \emph{2020 IEEE International Solid- State Circuits Conference -
  (ISSCC)}, pages 112--114, 2020.
\newblock \doi{10.1109/ISSCC19947.2020.9063149}.

\bibitem[Fossum(2005)]{fossum2005sub}
E.~R. Fossum.
\newblock What to do with sub-diffraction-limit {(SDL)} pixels?—a proposal
  for a gigapixel digital film sensor {(DFS)}.
\newblock In \emph{IEEE Workshop on Charge-Coupled Devices and Advanced Image
  Sensors}, pages 214--217, 2005.

\bibitem[Fossum(2011)]{Fossum:11}
E.~R. Fossum.
\newblock The quanta image sensor {(QIS)}: Concepts and challenges.
\newblock In \emph{Imaging and Applied Optics}, page JTuE1. Optica Publishing
  Group, 2011.
\newblock \doi{10.1364/COSI.2011.JTuE1}.
\newblock URL \url{http://opg.optica.org/abstract.cfm?URI=COSI-2011-JTuE1}.

\bibitem[Fossum et~al.(2016)Fossum, Ma, Masoodian, Anzagira, and
  Zizza]{fossum2016quanta}
E.~R. Fossum, J.~Ma, S.~Masoodian, L.~Anzagira, and R.~Zizza.
\newblock The quanta image sensor: Every photon counts.
\newblock \emph{Sensors}, 16\penalty0 (8):\penalty0 1260, 2016.

\bibitem[Gehrig et~al.(2020)Gehrig, Rebecq, Gallego, and
  Scaramuzza]{gehrig2020eklt}
D.~Gehrig, H.~Rebecq, G.~Gallego, and D.~Scaramuzza.
\newblock Eklt: Asynchronous photometric feature tracking using events and
  frames.
\newblock \emph{International Journal of Computer Vision}, 128\penalty0
  (3):\penalty0 601--618, 2020.

\bibitem[Graca and Delbruck(2021)]{graca2021unraveling}
R.~Graca and T.~Delbruck.
\newblock Unraveling the paradox of intensity-dependent {DVS} pixel noise.
\newblock \emph{arXiv preprint arXiv:2109.08640}, 2021.

\bibitem[Gupta et~al.(2010)Gupta, Agrawal, Veeraraghavan, and
  Narasimhan]{gupta2010flexible}
M.~Gupta, A.~Agrawal, A.~Veeraraghavan, and S.~G. Narasimhan.
\newblock Flexible voxels for motion-aware videography.
\newblock In \emph{Computer Vision--ECCV 2010: 11th European Conference on
  Computer Vision, Heraklion, Crete, Greece, September 5-11, 2010, Proceedings,
  Part I 11}, pages 100--114. Springer, 2010.

\bibitem[Gutierrez-Barragan et~al.(2022)Gutierrez-Barragan, Ingle, Seets,
  Gupta, and Velten]{Gutierrez-Barragan_2022_CVPR}
F.~Gutierrez-Barragan, A.~Ingle, T.~Seets, M.~Gupta, and A.~Velten.
\newblock Compressive single-photon {3D} cameras.
\newblock In \emph{Proceedings of the IEEE/CVF Conference on Computer Vision
  and Pattern Recognition (CVPR)}, pages 17854--17864, June 2022.

\bibitem[Gyongy et~al.(2018)Gyongy, Dutton, and Henderson]{gyongy2018single}
I.~Gyongy, N.~A. Dutton, and R.~K. Henderson.
\newblock Single-photon tracking for high-speed vision.
\newblock \emph{Sensors}, 18\penalty0 (2):\penalty0 323, 2018.

\bibitem[Gyongy et~al.(2020)Gyongy, Hutchings, Halimi, Tyler, Chan, Zhu,
  McLaughlin, Henderson, and Leach]{gyongy2020high}
I.~Gyongy, S.~W. Hutchings, A.~Halimi, M.~Tyler, S.~Chan, F.~Zhu,
  S.~McLaughlin, R.~K. Henderson, and J.~Leach.
\newblock High-speed {3D} sensing via hybrid-mode imaging and guided
  upsampling.
\newblock \emph{Optica}, 7\penalty0 (10):\penalty0 1253--1260, 2020.

\bibitem[Harris et~al.(1988)Harris, Stephens, et~al.]{harris1988combined}
C.~Harris, M.~Stephens, et~al.
\newblock A combined corner and edge detector.
\newblock In \emph{Alvey vision conference}, volume~15, pages 10--5244.
  Citeseer, 1988.

\bibitem[Hidalgo-Carri\'o et~al.(2022)Hidalgo-Carri\'o, Gallego, and
  Scaramuzza]{Hidalgo-Carrio_2022_CVPR}
J.~Hidalgo-Carri\'o, G.~Gallego, and D.~Scaramuzza.
\newblock Event-aided direct sparse odometry.
\newblock In \emph{Proceedings of the IEEE/CVF Conference on Computer Vision
  and Pattern Recognition (CVPR)}, pages 5781--5790, June 2022.

\bibitem[Horn and Schunck(1981)]{horn1981determining}
B.~K. Horn and B.~G. Schunck.
\newblock Determining optical flow.
\newblock \emph{Artificial intelligence}, 17\penalty0 (1-3):\penalty0 185--203,
  1981.

\bibitem[Hu et~al.(2021)Hu, Liu, and Delbruck]{Hu_2021_CVPR}
Y.~Hu, S.-C. Liu, and T.~Delbruck.
\newblock v2e: From video frames to realistic {DVS} events.
\newblock In \emph{Proceedings of the IEEE/CVF Conference on Computer Vision
  and Pattern Recognition (CVPR) Workshops}, pages 1312--1321, June 2021.

\bibitem[Igual(2019)]{igual2019photographic}
J.~Igual.
\newblock Photographic noise performance measures based on raw files analysis
  of consumer cameras.
\newblock \emph{Electronics}, 8\penalty0 (11):\penalty0 1284, 2019.

\bibitem[Ingle et~al.(2019)Ingle, Velten, and Gupta]{ingleHighFluxPassive2019}
A.~Ingle, A.~Velten, and M.~Gupta.
\newblock High {{Flux Passive Imaging With Single}}-{{Photon Sensors}}.
\newblock In \emph{Proceedings of the IEEE/CVF Conference on Computer Vision
  and Pattern Recognition (CVPR)}, June 2019.

\bibitem[Ingle et~al.(2021)Ingle, Seets, Buttafava, Gupta, Tosi, Gupta, and
  Velten]{inglePassiveInterPhotonImaging2021}
A.~Ingle, T.~Seets, M.~Buttafava, S.~Gupta, A.~Tosi, M.~Gupta, and A.~Velten.
\newblock Passive inter-photon imaging.
\newblock In \emph{Proceedings of the IEEE/CVF Conference on Computer Vision
  and Pattern Recognition (CVPR)}, June 2021.

\bibitem[Iwabuchi et~al.(2021)Iwabuchi, Kameda, and Hamamoto]{Iwabuchi2021}
K.~Iwabuchi, Y.~Kameda, and T.~Hamamoto.
\newblock Image quality improvements based on motion-based deblurring for
  single-photon imaging.
\newblock \emph{IEEE Access}, 9:\penalty0 30080--30094, 2021.
\newblock \doi{10.1109/ACCESS.2021.3059293}.

\bibitem[Jiang et~al.(2021)Jiang, Choi, Jiang, and Gu]{jiang2021hdr}
Y.~Jiang, I.~Choi, J.~Jiang, and J.~Gu.
\newblock {HDR} video reconstruction with tri-exposure quad-bayer sensors.
\newblock \emph{arXiv preprint arXiv:2103.10982}, 2021.

\bibitem[Levin et~al.(2008)Levin, Sand, Cho, Durand, and
  Freeman]{levin2008motion}
A.~Levin, P.~Sand, T.~S. Cho, F.~Durand, and W.~T. Freeman.
\newblock Motion-invariant photography.
\newblock \emph{ACM Transactions on Graphics (TOG)}, 27\penalty0 (3):\penalty0
  1--9, 2008.

\bibitem[Levoy and Hanrahan(1996)]{levoy1996light}
M.~Levoy and P.~Hanrahan.
\newblock Light field rendering.
\newblock In \emph{Proceedings of the 23rd annual conference on Computer
  graphics and interactive techniques}, pages 31--42, 1996.

\bibitem[Li et~al.(2020)Li, Qi, Gulve, Wei, Genov, Kutulakos, and
  Heidrich]{YuqiAndresonAcc2020}
Y.~Li, M.~Qi, R.~Gulve, M.~Wei, R.~Genov, K.~N. Kutulakos, and W.~Heidrich.
\newblock End-to-end video compressive sensing using anderson-accelerated
  unrolled networks.
\newblock In \emph{2020 IEEE International Conference on Computational
  Photography (ICCP)}, pages 1--12, 2020.
\newblock \doi{10.1109/ICCP48838.2020.9105237}.

\bibitem[Lichtsteiner(2003)]{lichtsteiner200364x64}
P.~Lichtsteiner.
\newblock {64x64 event-driven logarithmic temporal derivative silicon retina}.
\newblock In \emph{Program 2003 IEEE Workshop on CCD and AIS}, 2003.

\bibitem[Liu et~al.(2022)Liu, Gutierrez-Barragan, Ingle, Gupta, and
  Velten]{Liu_2022_WACV}
Y.~Liu, F.~Gutierrez-Barragan, A.~Ingle, M.~Gupta, and A.~Velten.
\newblock Single-photon camera guided extreme dynamic range imaging.
\newblock In \emph{Proceedings of the IEEE/CVF Winter Conference on
  Applications of Computer Vision (WACV)}, pages 1575--1585, January 2022.

\bibitem[Llull et~al.(2013)Llull, Liao, Yuan, Yang, Kittle, Carin, Sapiro, and
  Brady]{CACTI_Llull:13}
P.~Llull, X.~Liao, X.~Yuan, J.~Yang, D.~Kittle, L.~Carin, G.~Sapiro, and D.~J.
  Brady.
\newblock Coded aperture compressive temporal imaging.
\newblock \emph{Opt. Express}, 21\penalty0 (9):\penalty0 10526--10545, May
  2013.
\newblock \doi{10.1364/OE.21.010526}.
\newblock URL \url{https://opg.optica.org/oe/abstract.cfm?URI=oe-21-9-10526}.

\bibitem[Lucas and Kanade(1981)]{lucas1981iterative}
B.~D. Lucas and T.~Kanade.
\newblock An iterative image registration technique with an application to
  stereo vision.
\newblock In \emph{IJCAI'81: 7th international joint conference on Artificial
  intelligence}, volume~2, pages 674--679, 1981.

\bibitem[Ma et~al.(2017)Ma, Masoodian, Starkey, and Fossum]{Ma:17}
J.~Ma, S.~Masoodian, D.~A. Starkey, and E.~R. Fossum.
\newblock Photon-number-resolving megapixel image sensor at room temperature
  without avalanche gain.
\newblock \emph{Optica}, 4\penalty0 (12):\penalty0 1474--1481, Dec 2017.
\newblock \doi{10.1364/OPTICA.4.001474}.
\newblock URL
  \url{http://www.osapublishing.org/optica/abstract.cfm?URI=optica-4-12-1474}.

\bibitem[Ma et~al.(2020)Ma, Gupta, Ulku, Bruschini, Charbon, and
  Gupta]{ma_quanta_2020}
S.~Ma, S.~Gupta, A.~C. Ulku, C.~Bruschini, E.~Charbon, and M.~Gupta.
\newblock Quanta burst photography.
\newblock \emph{ACM Transactions on Graphics}, 39\penalty0 (4):\penalty0 1--16,
  July 2020.
\newblock ISSN 0730-0301, 1557-7368.

\bibitem[Ma et~al.(2023)Ma, Mos, Charbon, and Gupta]{Ma_2023_WACV}
S.~Ma, P.~Mos, E.~Charbon, and M.~Gupta.
\newblock Burst vision using single-photon cameras.
\newblock In \emph{Proceedings of the IEEE/CVF Winter Conference on
  Applications of Computer Vision (WACV)}, pages 5375--5385, January 2023.

\bibitem[Mann and Picard(1994)]{mann1994beingundigital}
S.~Mann and R.~Picard.
\newblock Beingundigital’with digital cameras.
\newblock \emph{MIT Media Lab Perceptual}, 1:\penalty0 2, 1994.

\bibitem[Martel et~al.(2020)Martel, Mueller, Carey, Dudek, and
  Wetzstein]{martel2020neural}
J.~N. Martel, L.~K. Mueller, S.~J. Carey, P.~Dudek, and G.~Wetzstein.
\newblock Neural sensors: Learning pixel exposures for {HDR} imaging and video
  compressive sensing with programmable sensors.
\newblock \emph{IEEE Transactions on Pattern Analysis and Machine
  Intelligence}, 42\penalty0 (7):\penalty0 1642--1653, 2020.

\bibitem[Martel et~al.(2016)Martel, Müller, Carey, and Dudek]{Martel2016}
J.~N.~P. Martel, L.~K. Müller, S.~J. Carey, and P.~Dudek.
\newblock Parallel {HDR} tone mapping and auto-focus on a cellular processor
  array vision chip.
\newblock In \emph{2016 IEEE International Symposium on Circuits and Systems
  (ISCAS)}, pages 1430--1433, 2016.
\newblock \doi{10.1109/ISCAS.2016.7527519}.

\bibitem[Martel et~al.(2017)Martel, Müller, Carey, and Dudek]{Martel2017}
J.~N.~P. Martel, L.~K. Müller, S.~J. Carey, and P.~Dudek.
\newblock High-speed depth from focus on a programmable vision chip using a
  focus tunable lens.
\newblock In \emph{2017 IEEE International Symposium on Circuits and Systems
  (ISCAS)}, pages 1--4, 2017.
\newblock \doi{10.1109/ISCAS.2017.8050548}.

\bibitem[Metzler et~al.(2020)Metzler, Ikoma, Peng, and
  Wetzstein]{Metzler_2020_CVPR}
C.~A. Metzler, H.~Ikoma, Y.~Peng, and G.~Wetzstein.
\newblock Deep optics for single-shot high-dynamic-range imaging.
\newblock In \emph{Proceedings of the IEEE/CVF Conference on Computer Vision
  and Pattern Recognition (CVPR)}, June 2020.

\bibitem[Morimoto et~al.(2020)Morimoto, Ardelean, Wu, Ulku, Antolovic,
  Bruschini, and Charbon]{morimoto_megapixel_2020}
K.~Morimoto, A.~Ardelean, M.-L. Wu, A.~C. Ulku, I.~M. Antolovic, C.~Bruschini,
  and E.~Charbon.
\newblock Megapixel time-gated {{SPAD}} image sensor for {{2D}} and {{3D}}
  imaging applications.
\newblock \emph{Optica}, 7\penalty0 (4):\penalty0 346--354, Apr. 2020.

\bibitem[Morimoto et~al.(2021)Morimoto, Iwata, Shinohara, Sekine, Abdelghafar,
  Tsuchiya, Kuroda, Tojima, Endo, Maehashi, Ota, Sasago, Maekawa, Hikosaka,
  Kanou, Kato, Tezuka, Yoshizaki, Ogawa, Uehira, Ehara, Inui, Matsuno, Sakurai,
  and Ichikawa]{Morimoto_cannon}
K.~Morimoto, J.~Iwata, M.~Shinohara, H.~Sekine, A.~Abdelghafar, H.~Tsuchiya,
  Y.~Kuroda, K.~Tojima, W.~Endo, Y.~Maehashi, Y.~Ota, T.~Sasago, S.~Maekawa,
  S.~Hikosaka, T.~Kanou, A.~Kato, T.~Tezuka, S.~Yoshizaki, T.~Ogawa, K.~Uehira,
  A.~Ehara, F.~Inui, Y.~Matsuno, K.~Sakurai, and T.~Ichikawa.
\newblock 3.2 megapixel {3D}-stacked charge focusing {SPAD} for low-light
  imaging and depth sensing.
\newblock In \emph{2021 IEEE International Electron Devices Meeting (IEDM)},
  pages 20.2.1--20.2.4, 2021.
\newblock \doi{10.1109/IEDM19574.2021.9720605}.

\bibitem[Namiki et~al.(2022)Namiki, Sato, Kameda, and
  Hamamoto]{namiki2022imaging}
S.~Namiki, S.~Sato, Y.~Kameda, and T.~Hamamoto.
\newblock Imaging method using multi-threshold pattern for photon detection of
  quanta image sensor.
\newblock In \emph{International Workshop on Advanced Imaging Technology
  (IWAIT) 2022}, volume 12177, page 1217702. SPIE, 2022.

\bibitem[Nayar et~al.(2006)Nayar, Branzoi, and Boult]{nayar2006programmable}
S.~K. Nayar, V.~Branzoi, and T.~E. Boult.
\newblock Programmable imaging: Towards a flexible camera.
\newblock \emph{International Journal of Computer Vision}, 70:\penalty0 7--22,
  2006.

\bibitem[Posch et~al.(2011)Posch, Matolin, and Wohlgenannt]{ATIS_2011}
C.~Posch, D.~Matolin, and R.~Wohlgenannt.
\newblock {A QVGA 143 dB Dynamic Range Frame-Free PWM Image Sensor With
  Lossless Pixel-Level Video Compression and Time-Domain CDS}.
\newblock \emph{IEEE Journal of Solid-State Circuits}, 46\penalty0
  (1):\penalty0 259--275, 2011.
\newblock \doi{10.1109/JSSC.2010.2085952}.

\bibitem[Raskar et~al.(2006)Raskar, Agrawal, and Tumblin]{raskar2006coded}
R.~Raskar, A.~Agrawal, and J.~Tumblin.
\newblock Coded exposure photography: motion deblurring using fluttered
  shutter.
\newblock In \emph{Acm Siggraph 2006 Papers}, pages 795--804. 2006.

\bibitem[Rebecq et~al.(2019)Rebecq, Ranftl, Koltun, and
  Scaramuzza]{Rebecq_2019_CVPR}
H.~Rebecq, R.~Ranftl, V.~Koltun, and D.~Scaramuzza.
\newblock Events-to-video: Bringing modern computer vision to event cameras.
\newblock In \emph{Proceedings of the IEEE/CVF Conference on Computer Vision
  and Pattern Recognition (CVPR)}, June 2019.

\bibitem[Reddy et~al.(2011)Reddy, Veeraraghavan, and Chellappa]{P2C2}
D.~Reddy, A.~Veeraraghavan, and R.~Chellappa.
\newblock {P2C2}: Programmable pixel compressive camera for high speed imaging.
\newblock In \emph{CVPR 2011}, pages 329--336, 2011.
\newblock \doi{10.1109/CVPR.2011.5995542}.

\bibitem[Ren et~al.(2018)Ren, Connolly, Halimi, Altmann, McLaughlin, Gyongy,
  Henderson, and Buller]{ren2018high}
X.~Ren, P.~W. Connolly, A.~Halimi, Y.~Altmann, S.~McLaughlin, I.~Gyongy, R.~K.
  Henderson, and G.~S. Buller.
\newblock High-resolution depth profiling using a range-gated {CMOS SPAD}
  quanta image sensor.
\newblock \emph{Optics express}, 26\penalty0 (5):\penalty0 5541--5557, 2018.

\bibitem[Scheerlinck et~al.(2020)Scheerlinck, Rebecq, Gehrig, Barnes, Mahony,
  and Scaramuzza]{Scheerlinck_2020_WACV}
C.~Scheerlinck, H.~Rebecq, D.~Gehrig, N.~Barnes, R.~Mahony, and D.~Scaramuzza.
\newblock Fast image reconstruction with an event camera.
\newblock In \emph{Proceedings of the IEEE/CVF Winter Conference on
  Applications of Computer Vision (WACV)}, March 2020.

\bibitem[Seets et~al.(2021)Seets, Ingle, Laurenzis, and
  Velten]{Seets_2021_WACV}
T.~Seets, A.~Ingle, M.~Laurenzis, and A.~Velten.
\newblock Motion adaptive deblurring with single-photon cameras.
\newblock In \emph{Proceedings of the IEEE/CVF Winter Conference on
  Applications of Computer Vision (WACV)}, pages 1945--1954, January 2021.

\bibitem[Seo et~al.(2017)Seo, Shirakawa, Masuda, Kawata, Kagawa, Yasutomi, and
  Kawahito]{Woong4bucket2017}
M.-W. Seo, Y.~Shirakawa, Y.~Masuda, Y.~Kawata, K.~Kagawa, K.~Yasutomi, and
  S.~Kawahito.
\newblock 4.3 a programmable sub-nanosecond time-gated 4-tap lock-in pixel cmos
  image sensor for real-time fluorescence lifetime imaging microscopy.
\newblock In \emph{2017 IEEE International Solid-State Circuits Conference
  (ISSCC)}, pages 70--71, 2017.
\newblock \doi{10.1109/ISSCC.2017.7870265}.

\bibitem[Serrano-Gotarredona and Linares-Barranco(2013)]{serrano2013128}
T.~Serrano-Gotarredona and B.~Linares-Barranco.
\newblock A 128 $\times$ 128 1.5\% contrast sensitivity 0.9\% {FPN} 3 $\mu$s
  latency 4 {mW} asynchronous frame-free dynamic vision sensor using
  transimpedance preamplifiers.
\newblock \emph{IEEE Journal of Solid-State Circuits}, 48\penalty0
  (3):\penalty0 827--838, 2013.

\bibitem[Shedligeri et~al.(2021)Shedligeri, S, and Mitra]{Shedligeri_2021_WACV}
P.~Shedligeri, A.~S, and K.~Mitra.
\newblock A unified framework for compressive video recovery from coded
  exposure techniques.
\newblock In \emph{Proceedings of the IEEE/CVF Winter Conference on
  Applications of Computer Vision (WACV)}, pages 1600--1609, January 2021.

\bibitem[Shi et~al.(2022)Shi, Song, Li, Li, Wei, Liu, and Jin]{shi2022review}
C.~Shi, N.~Song, W.~Li, Y.~Li, B.~Wei, H.~Liu, and J.~Jin.
\newblock A review of event-based indoor positioning and navigation.
\newblock 2022.

\bibitem[Sitzmann et~al.(2018)Sitzmann, Diamond, Peng, Dun, Boyd, Heidrich,
  Heide, and Wetzstein]{sitzmann2018end}
V.~Sitzmann, S.~Diamond, Y.~Peng, X.~Dun, S.~Boyd, W.~Heidrich, F.~Heide, and
  G.~Wetzstein.
\newblock End-to-end optimization of optics and image processing for achromatic
  extended depth of field and super-resolution imaging.
\newblock \emph{ACM Transactions on Graphics (TOG)}, 37\penalty0 (4):\penalty0
  114, 2018.

\bibitem[Taverni et~al.(2018)Taverni, Paul~Moeys, Li, Cavaco, Motsnyi, San
  Segundo~Bello, and Delbruck]{GemmacolorDAVIS}
G.~Taverni, D.~Paul~Moeys, C.~Li, C.~Cavaco, V.~Motsnyi, D.~San Segundo~Bello,
  and T.~Delbruck.
\newblock Front and back illuminated dynamic and active pixel vision sensors
  comparison.
\newblock \emph{IEEE Transactions on Circuits and Systems II: Express Briefs},
  65\penalty0 (5):\penalty0 677--681, 2018.
\newblock \doi{10.1109/TCSII.2018.2824899}.

\bibitem[Teed and Deng(2020)]{teed2020raft}
Z.~Teed and J.~Deng.
\newblock Raft: Recurrent all-pairs field transforms for optical flow.
\newblock In \emph{European Conference on Computer Vision}, 2020.

\bibitem[Tulyakov et~al.(2021)Tulyakov, Gehrig, Georgoulis, Erbach, Gehrig, Li,
  and Scaramuzza]{Tulyakov_2021_CVPR}
S.~Tulyakov, D.~Gehrig, S.~Georgoulis, J.~Erbach, M.~Gehrig, Y.~Li, and
  D.~Scaramuzza.
\newblock Time lens: Event-based video frame interpolation.
\newblock In \emph{Proceedings of the IEEE/CVF Conference on Computer Vision
  and Pattern Recognition (CVPR)}, pages 16155--16164, June 2021.

\bibitem[Ulku et~al.(2019)Ulku, Bruschini, Antolovic, Kuo, Ankri, Weiss,
  Michalet, and Charbon]{ulku512512SPAD2019}
A.~C. Ulku, C.~Bruschini, I.~M. Antolovic, Y.~Kuo, R.~Ankri, S.~Weiss,
  X.~Michalet, and E.~Charbon.
\newblock A 512 \texttimes{} 512 {{SPAD Image Sensor With Integrated Gating}}
  for {{Widefield FLIM}}.
\newblock \emph{IEEE Journal of Selected Topics in Quantum Electronics},
  25\penalty0 (1):\penalty0 1--12, Jan. 2019.
\newblock ISSN 1077-260X, 1558-4542.
\newblock \doi{10.1109/JSTQE.2018.2867439}.

\bibitem[Wan et~al.(2012)Wan, Li, Agranov, Levoy, and
  Horowitz]{Wan2012MultiBucket}
G.~Wan, X.~Li, G.~Agranov, M.~Levoy, and M.~Horowitz.
\newblock {CMOS} image sensors with multi-bucket pixels for computational
  photography.
\newblock \emph{IEEE Journal of Solid-State Circuits}, 47\penalty0
  (4):\penalty0 1031--1042, 2012.
\newblock \doi{10.1109/JSSC.2012.2185189}.

\bibitem[Wei et~al.(2018)Wei, Sarhangnejad, Xia, Gusev, Katic, Genov, and
  Kutulakos]{Wei_2018_ECCV}
M.~Wei, N.~Sarhangnejad, Z.~Xia, N.~Gusev, N.~Katic, R.~Genov, and K.~N.
  Kutulakos.
\newblock Coded two-bucket cameras for computer vision.
\newblock In \emph{Proceedings of the European Conference on Computer Vision
  (ECCV)}, September 2018.

\bibitem[White et~al.(2022)White, Ghajari, Zhang, Dave, Veeraraghavan, and
  Molnar]{White2022DSPAD}
M.~White, S.~Ghajari, T.~Zhang, A.~Dave, A.~Veeraraghavan, and A.~Molnar.
\newblock A differential {SPAD} array architecture in {0.18 $\mu$m CMOS for
  HDR} imaging.
\newblock In \emph{2022 IEEE International Symposium on Circuits and Systems
  (ISCAS)}, pages 292--296, 2022.
\newblock \doi{10.1109/ISCAS48785.2022.9937558}.

\bibitem[Yamazaki et~al.(2017)Yamazaki, Katayama, Uehara, Nose, Kobayashi,
  Shida, Odahara, Takamiya, Hisamatsu, Matsumoto, Miyashita, Watanabe, Izawa,
  Muramatsu, and Ishikawa]{Tomohiro2017}
T.~Yamazaki, H.~Katayama, S.~Uehara, A.~Nose, M.~Kobayashi, S.~Shida,
  M.~Odahara, K.~Takamiya, Y.~Hisamatsu, S.~Matsumoto, L.~Miyashita,
  Y.~Watanabe, T.~Izawa, Y.~Muramatsu, and M.~Ishikawa.
\newblock 4.9 a 1ms high-speed vision chip with {3D}-stacked {140GOPS}
  column-parallel {PEs} for spatio-temporal image processing.
\newblock In \emph{2017 IEEE International Solid-State Circuits Conference
  (ISSCC)}, pages 82--83, 2017.
\newblock \doi{10.1109/ISSCC.2017.7870271}.

\bibitem[Yang et~al.(2012)Yang, Lu, Sbaiz, and Vetterli]{yang_poisson_model}
F.~Yang, Y.~M. Lu, L.~Sbaiz, and M.~Vetterli.
\newblock Bits from photons: Oversampled image acquisition using binary poisson
  statistics.
\newblock \emph{IEEE Transactions on Image Processing}, 21\penalty0
  (4):\penalty0 1421--1436, 2012.
\newblock \doi{10.1109/TIP.2011.2179306}.

\bibitem[Yuan et~al.(2022)Yuan, Liu, Suo, Durand, and Dai]{Yuan2022PnPADMM}
X.~Yuan, Y.~Liu, J.~Suo, F.~Durand, and Q.~Dai.
\newblock Plug-and-play algorithms for video snapshot compressive imaging.
\newblock \emph{IEEE Transactions on Pattern Analysis and Machine
  Intelligence}, 44\penalty0 (10):\penalty0 7093--7111, 2022.
\newblock \doi{10.1109/TPAMI.2021.3099035}.

\bibitem[Zhang et~al.(2021)Zhang, Yang, Fu, Wei, Yin, and
  Dong]{zhang2021object}
J.~Zhang, X.~Yang, Y.~Fu, X.~Wei, B.~Yin, and B.~Dong.
\newblock Object tracking by jointly exploiting frame and event domain.
\newblock In \emph{Proceedings of the IEEE/CVF International Conference on
  Computer Vision (ICCV)}, pages 13043--13052, 2021.

\bibitem[Zhang et~al.(2020)Zhang, Li, Zuo, Zhang, Van~Gool, and
  Timofte]{zhang2020plug}
K.~Zhang, Y.~Li, W.~Zuo, L.~Zhang, L.~Van~Gool, and R.~Timofte.
\newblock Plug-and-play image restoration with deep denoiser prior.
\newblock \emph{arXiv preprint}, 2020.

\bibitem[Zhang et~al.(2018)Zhang, Isola, Efros, Shechtman, and
  Wang]{Zhang_2018_CVPR}
R.~Zhang, P.~Isola, A.~A. Efros, E.~Shechtman, and O.~Wang.
\newblock The unreasonable effectiveness of deep features as a perceptual
  metric.
\newblock In \emph{Proceedings of the IEEE Conference on Computer Vision and
  Pattern Recognition (CVPR)}, June 2018.

\bibitem[Zhang et~al.(2022)Zhang, White, Dave, Ghajari, Raghuram, Molnar, and
  Veeraraghavan]{Zhang2022First}
T.~Zhang, M.~J. White, A.~Dave, S.~Ghajari, A.~Raghuram, A.~C. Molnar, and
  A.~Veeraraghavan.
\newblock First arrival differential {LiDAR}.
\newblock In \emph{2022 IEEE International Conference on Computational
  Photography (ICCP)}, pages 1--12, 2022.
\newblock \doi{10.1109/ICCP54855.2022.9887683}.

\bibitem[Zhang and Lin(2017)]{zhang2017light}
W.~Zhang and L.~Lin.
\newblock Light field flow estimation based on occlusion detection.
\newblock \emph{Journal of Computer and Communications}, 5\penalty0
  (3):\penalty0 1--9, 2017.

\end{thebibliography}

\begin{thebibliography}{27}
\providecommand{\natexlab}[1]{#1}
\providecommand{\url}[1]{\texttt{#1}}
\expandafter\ifx\csname urlstyle\endcsname\relax
  \providecommand{\doi}[1]{doi: #1}\else
  \providecommand{\doi}{doi: \begingroup \urlstyle{rm}\Url}\fi

\bibitem[Ardelean(2023)]{ardelean2023computational}
A.~Ardelean.
\newblock \emph{Computational Imaging SPAD Cameras}.
\newblock PhD thesis, École polytechnique fédérale de Lausanne, 2023.

\bibitem[Baker et~al.(2011)Baker, Scharstein, Lewis, Roth, Black, and
  Szeliski]{baker2011database}
S.~Baker, D.~Scharstein, J.~Lewis, S.~Roth, M.~J. Black, and R.~Szeliski.
\newblock A database and evaluation methodology for optical flow.
\newblock \emph{International journal of computer vision}, 92:\penalty0 1--31,
  2011.

\bibitem[Brandli et~al.(2014)Brandli, Berner, Yang, Liu, and
  Delbruck]{DAVIS240}
C.~Brandli, R.~Berner, M.~Yang, S.-C. Liu, and T.~Delbruck.
\newblock A 240 × 180 {130 dB 3 µs} latency global shutter spatiotemporal
  vision sensor.
\newblock \emph{IEEE Journal of Solid-State Circuits}, 49\penalty0
  (10):\penalty0 2333--2341, 2014.
\newblock \doi{10.1109/JSSC.2014.2342715}.

\bibitem[Bresenham(1965)]{bresenham1965algorithm}
J.~E. Bresenham.
\newblock Algorithm for computer control of a digital plotter.
\newblock \emph{IBM Systems journal}, 4\penalty0 (1):\penalty0 25--30, 1965.

\bibitem[Chan et~al.(2017)Chan, Wang, and Elgendy]{chan2017PnP}
S.~H. Chan, X.~Wang, and O.~A. Elgendy.
\newblock Plug-and-play admm for image restoration: Fixed-point convergence and
  applications.
\newblock \emph{IEEE Transactions on Computational Imaging}, 3\penalty0
  (1):\penalty0 84--98, 2017.
\newblock \doi{10.1109/TCI.2016.2629286}.

\bibitem[Chen and Guo(2019)]{Chen_2019_CVPR_Workshops}
S.~Chen and M.~Guo.
\newblock Live demonstration: Celex-v: A 1m pixel multi-mode event-based
  sensor.
\newblock In \emph{Proceedings of the IEEE/CVF Conference on Computer Vision
  and Pattern Recognition (CVPR) Workshops}, June 2019.

\bibitem[Chengshuai~Yang(2022)]{Yuan2022ELPUnfolding}
X.~Y. Chengshuai~Yang, Shiyu~Zhang.
\newblock Ensemble learning priors driven deep unfolding for scalable video
  snapshot compressive imaging.
\newblock In \emph{IEEE European Conference on Computer Vision (ECCV)}, 2022.

\bibitem[Gallego et~al.(2018)Gallego, Rebecq, and
  Scaramuzza]{Gallego_2018_CVPR}
G.~Gallego, H.~Rebecq, and D.~Scaramuzza.
\newblock A unifying contrast maximization framework for event cameras, with
  applications to motion, depth, and optical flow estimation.
\newblock In \emph{Proceedings of the IEEE Conference on Computer Vision and
  Pattern Recognition (CVPR)}, June 2018.

\bibitem[Li et~al.(2020)Li, Qi, Gulve, Wei, Genov, Kutulakos, and
  Heidrich]{YuqiAndresonAcc2020}
Y.~Li, M.~Qi, R.~Gulve, M.~Wei, R.~Genov, K.~N. Kutulakos, and W.~Heidrich.
\newblock End-to-end video compressive sensing using anderson-accelerated
  unrolled networks.
\newblock In \emph{2020 IEEE International Conference on Computational
  Photography (ICCP)}, pages 1--12, 2020.
\newblock \doi{10.1109/ICCP48838.2020.9105237}.

\bibitem[Liu et~al.(2019)Liu, Yuan, Suo, Brady, and Dai]{Liu19DeSCI}
Y.~Liu, X.~Yuan, J.~Suo, D.~J. Brady, and Q.~Dai.
\newblock Rank minimization for snapshot compressive imaging.
\newblock \emph{IEEE Trans. Pattern Anal. Mach. Intell.}, 41\penalty0
  (12):\penalty0 2990 -- 3006, 2019.
\newblock \doi{10.1109/TPAMI.2018.2873587}.
\newblock URL \url{https://doi.org/10.1109/TPAMI.2018.2873587}.

\bibitem[Ma et~al.(2020)Ma, Gupta, Ulku, Bruschini, Charbon, and
  Gupta]{ma_quanta_2020}
S.~Ma, S.~Gupta, A.~C. Ulku, C.~Bruschini, E.~Charbon, and M.~Gupta.
\newblock Quanta burst photography.
\newblock \emph{ACM Transactions on Graphics}, 39\penalty0 (4):\penalty0 1--16,
  July 2020.
\newblock ISSN 0730-0301, 1557-7368.

\bibitem[Maqueda et~al.(2018)Maqueda, Loquercio, Gallego, García, and
  Scaramuzza]{Maqueda_2018_CVPR}
A.~I. Maqueda, A.~Loquercio, G.~Gallego, N.~García, and D.~Scaramuzza.
\newblock Event-based vision meets deep learning on steering prediction for
  self-driving cars.
\newblock In \emph{Proceedings of the IEEE Conference on Computer Vision and
  Pattern Recognition (CVPR)}, June 2018.

\bibitem[Rebecq et~al.(2019)Rebecq, Ranftl, Koltun, and
  Scaramuzza]{Rebecq_2019_CVPR}
H.~Rebecq, R.~Ranftl, V.~Koltun, and D.~Scaramuzza.
\newblock Events-to-video: Bringing modern computer vision to event cameras.
\newblock In \emph{Proceedings of the IEEE/CVF Conference on Computer Vision
  and Pattern Recognition (CVPR)}, June 2019.

\bibitem[Shedligeri et~al.(2021)Shedligeri, S, and Mitra]{Shedligeri_2021_WACV}
P.~Shedligeri, A.~S, and K.~Mitra.
\newblock A unified framework for compressive video recovery from coded
  exposure techniques.
\newblock In \emph{Proceedings of the IEEE/CVF Winter Conference on
  Applications of Computer Vision (WACV)}, pages 1600--1609, January 2021.

\bibitem[Snoeij et~al.(2007)Snoeij, Theuwissen, Makinwa, and
  Huijsing]{SnoeijJSSC2007}
M.~F. Snoeij, A.~J.~P. Theuwissen, K.~A.~A. Makinwa, and J.~H. Huijsing.
\newblock {Multiple-Ramp Column-Parallel ADC Architectures for CMOS Image
  Sensors}.
\newblock \emph{IEEE JSSC}, 2007.
\newblock \doi{10.1109/JSSC.2007.908720}.

\bibitem[Tassano et~al.(2020)Tassano, Delon, and Veit]{Tassano_2020_CVPR}
M.~Tassano, J.~Delon, and T.~Veit.
\newblock Fastdvdnet: Towards real-time deep video denoising without flow
  estimation.
\newblock In \emph{Proceedings of the IEEE/CVF Conference on Computer Vision
  and Pattern Recognition (CVPR)}, June 2020.

\bibitem[Teed and Deng(2020)]{teed2020raft}
Z.~Teed and J.~Deng.
\newblock Raft: Recurrent all-pairs field transforms for optical flow.
\newblock In \emph{European Conference on Computer Vision}, 2020.

\bibitem[Telea(2004)]{telea2004image}
A.~Telea.
\newblock An image inpainting technique based on the fast marching method.
\newblock \emph{Journal of graphics tools}, 9\penalty0 (1):\penalty0 23--34,
  2004.

\bibitem[Ulku et~al.(2019)Ulku, Bruschini, Antolovic, Kuo, Ankri, Weiss,
  Michalet, and Charbon]{ulku512512SPAD2019}
A.~C. Ulku, C.~Bruschini, I.~M. Antolovic, Y.~Kuo, R.~Ankri, S.~Weiss,
  X.~Michalet, and E.~Charbon.
\newblock A 512 \texttimes{} 512 {{SPAD Image Sensor With Integrated Gating}}
  for {{Widefield FLIM}}.
\newblock \emph{IEEE Journal of Selected Topics in Quantum Electronics},
  25\penalty0 (1):\penalty0 1--12, Jan. 2019.
\newblock ISSN 1077-260X, 1558-4542.
\newblock \doi{10.1109/JSTQE.2018.2867439}.

\bibitem[Venkatakrishnan et~al.(2013)Venkatakrishnan, Bouman, and
  Wohlberg]{venkatakrishnan2013PnP}
S.~V. Venkatakrishnan, C.~A. Bouman, and B.~Wohlberg.
\newblock Plug-and-play priors for model based reconstruction.
\newblock In \emph{2013 IEEE Global Conference on Signal and Information
  Processing}, pages 945--948, 2013.
\newblock \doi{10.1109/GlobalSIP.2013.6737048}.

\bibitem[Wan et~al.(2022)Wan, Dai, and Mao]{DCEIFlow2022}
Z.~Wan, Y.~Dai, and Y.~Mao.
\newblock Learning dense and continuous optical flow from an event camera.
\newblock \emph{IEEE Transactions on Image Processing}, 31:\penalty0
  7237--7251, 2022.
\newblock \doi{10.1109/TIP.2022.3220938}.

\bibitem[Wang et~al.(2021)Wang, Zhang, Cheng, Chen, and Yuan]{Wang_2021_CVPR}
Z.~Wang, H.~Zhang, Z.~Cheng, B.~Chen, and X.~Yuan.
\newblock Metasci: Scalable and adaptive reconstruction for video compressive
  sensing.
\newblock In \emph{Proceedings of the IEEE/CVF Conference on Computer Vision
  and Pattern Recognition (CVPR)}, pages 2083--2092, June 2021.

\bibitem[Wu et~al.(2021)Wu, Zhang, and Mou]{Wu_2021_ICCV}
Z.~Wu, J.~Zhang, and C.~Mou.
\newblock Dense deep unfolding network with {3D-CNN} prior for snapshot
  compressive imaging.
\newblock In \emph{Proceedings of the IEEE/CVF International Conference on
  Computer Vision (ICCV)}, pages 4892--4901, October 2021.

\bibitem[Yuan(2016)]{gapTVYuan2016}
X.~Yuan.
\newblock Generalized alternating projection based total variation minimization
  for compressive sensing.
\newblock In \emph{2016 IEEE International Conference on Image Processing
  (ICIP)}, pages 2539--2543, 2016.
\newblock \doi{10.1109/ICIP.2016.7532817}.

\bibitem[Yuan et~al.(2021)Yuan, Brady, and Katsaggelos]{yuan2021snapshot}
X.~Yuan, D.~J. Brady, and A.~K. Katsaggelos.
\newblock Snapshot compressive imaging: Theory, algorithms, and applications.
\newblock \emph{IEEE Signal Processing Magazine}, 38\penalty0 (2):\penalty0
  65--88, 2021.

\bibitem[Yuan et~al.(2022)Yuan, Liu, Suo, Durand, and Dai]{Yuan2022PnPADMM}
X.~Yuan, Y.~Liu, J.~Suo, F.~Durand, and Q.~Dai.
\newblock Plug-and-play algorithms for video snapshot compressive imaging.
\newblock \emph{IEEE Transactions on Pattern Analysis and Machine
  Intelligence}, 44\penalty0 (10):\penalty0 7093--7111, 2022.
\newblock \doi{10.1109/TPAMI.2021.3099035}.

\bibitem[Zhu et~al.(2019)Zhu, Yuan, Chaney, and
  Daniilidis]{zhu2019unsupervised}
A.~Z. Zhu, L.~Yuan, K.~Chaney, and K.~Daniilidis.
\newblock Unsupervised event-based learning of optical flow, depth, and
  egomotion.
\newblock In \emph{Proceedings of the IEEE/CVF Conference on Computer Vision
  and Pattern Recognition}, pages 989--997, 2019.

\end{thebibliography}
